\documentclass[10pt,journal,compsoc]{IEEEtran}
\ifCLASSOPTIONcompsoc
  \usepackage[nocompress]{cite}
\else
  \usepackage{cite}
\fi
\ifCLASSINFOpdf
\else
\fi
\usepackage{amsmath}
\usepackage{graphicx}
\usepackage{subfigure}
\usepackage{booktabs}
\usepackage{amssymb}
\usepackage{subfigure}
\usepackage{tabu}
\usepackage{multirow}
\usepackage{algorithm}
\usepackage{algorithmic}
\usepackage[capitalise]{cleveref}
\usepackage{cite}

\begin{document}
%
\title{Unsupervised Domain Adaptation for Image Classification via Structure-Conditioned Adversarial Learning}
%
%
%
%

\author{Hui~Wang,
	Jian~Tian,
	Songyuan~Li,
	Hanbin~Zhao,
	Qi Tian,
	Fei Wu,
	and~Xi~Li
	\IEEEcompsocitemizethanks{\IEEEcompsocthanksitem H.~Wang, J.~Tian, S.~Li, H.~Zhao, F. Wu, and X.~Li are with College of Computer Science and Technology, Zhejiang University, Hangzhou 310027, China. \protect
	
	E-mail: \{wanghui\_17, tianjian29, leizungjyun, zhaohanbin, wufei, xilizju\}@zju.edu.cn. \protect
	\IEEEcompsocthanksitem Q.~Tian is with Cloud BU, Huawei Technologies, Shenzhen 518129, China. \protect
	
	E-mail: tian.qi1@huawei.com. \protect
}
	\thanks{(Corresponding author: Xi Li.)}
}
%
%

\markboth{IEEE TRANSACTIONS ON PATTERN ANALYSIS AND MACHINE INTELLIGENCE}%
{Shell \MakeLowercase{\textit{et al.}}: Bare Demo of IEEEtran.cls for Computer Society Journals}
%



\IEEEtitleabstractindextext{%
\begin{abstract}
	Unsupervised domain adaptation (UDA) typically carries out knowledge transfer from a label-rich source domain to an unlabeled target domain by adversarial learning. In principle, existing UDA approaches mainly focus on the global distribution alignment between domains while ignoring the intrinsic local distribution properties. Motivated by this observation, we propose an end-to-end structure-conditioned adversarial learning scheme (SCAL) that is able to preserve the intra-class compactness during domain distribution alignment. By using local structures as structure-aware conditions, the proposed scheme is implemented in a structure-conditioned adversarial learning pipeline. The above learning procedure is iteratively performed by alternating between local structures establishment and structure-conditioned adversarial learning. Experimental results demonstrate the effectiveness of the proposed scheme in UDA scenarios.
\end{abstract}

\begin{IEEEkeywords}
Unsupervised Domain Adaptation, Image Classification, Adversarial Learning, Clustering.
\end{IEEEkeywords}}

\maketitle

\IEEEdisplaynontitleabstractindextext

%
\IEEEpeerreviewmaketitle

\IEEEraisesectionheading{\section{Introduction}\label{sec:introduction}}

\begin{figure}[t]
	\centering
	\includegraphics[width = 1\columnwidth]{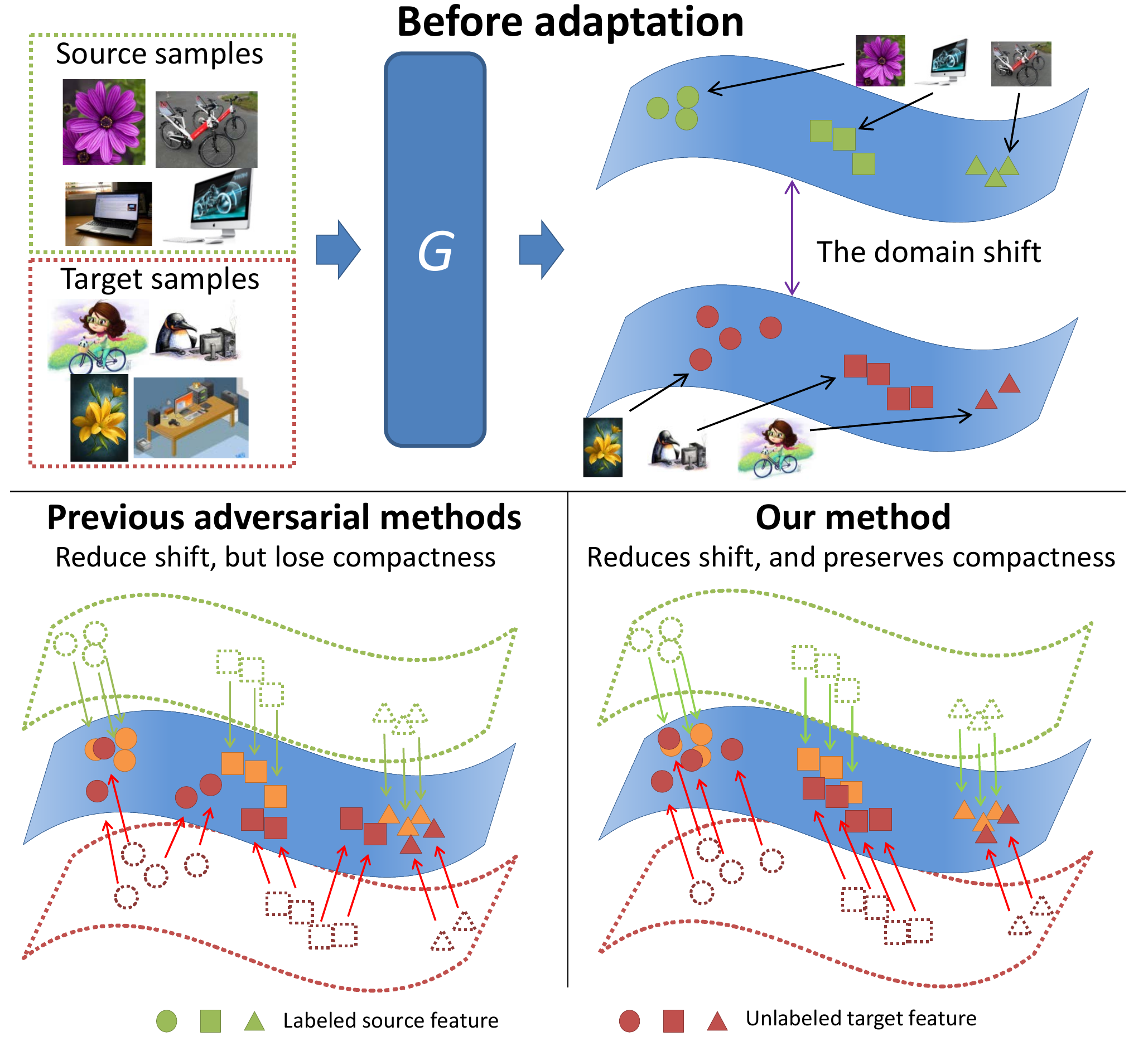}	
	\caption{(Best viewed in color.) Comparison of different UDA methods. The $G$ denotes the feature extractor and the shape of each feature refers to its affiliated class. \textbf{Top:} Before adaptation, the source domain features and the target domain features belong to two different manifolds. \textbf{Bottom-left:} Previous adversarial method tries to reduce the domain shift but fails to keep the local structures of the target domain. \textbf{Bottom-right:} Our method manages to preserve the target local structures and reduce the domain shift as well.}
		\label{fig:disca}
\end{figure}

\IEEEPARstart{R}{ecent} years have witnessed a great development of unsupervised domain adaptation (UDA), which aims to transfer the knowledge of
a label-rich source domain to an unlabeled target domain (with the same class information as the source domain). 
However, as shown in~\cref{fig:disca}, the features of the source domain and the target domain are often distributed on different manifolds, which poses the problem of adapting a well-trained model from the source domain to the target domain~\cite{ben2007analysis,ben2010theory,torralba2011unbiased}.

In the literature, a typical solution to UDA is based on adversarial domain
adaptation~\cite{ganin2014unsupervised,ganin2016domain,long2018conditional,tzeng2017adversarial},
which seeks to bridge the domain gap across domains by
playing an adversarial game between two players, i.e., a feature extractor and a domain classifier.
The feature extractor contrives to fool the domain classifier. Conversely, the domain classifier strives to determine whether the extracted features belong to the source or target domain.
As a result, such a domain alignment scheme \emph{globally} narrows down the feature distribution gap between domains, without \emph{locally} modeling the intra-class compactness of the target domain. 
Therefore, these methods are likely to have some difficulties in well capturing the consistency of the local structures for the target domain features, as illustrated in~\cref{fig:disca}.


Motivated by the above observations, we propose to learn the local structures of the target domain and integrate the structures into an adversarial learning process, resulting in a framework called structure-conditioned adversarial learning (SCAL). 
The key to SCAL lies in the novel clustering-based conditions which contain the local structures of target domain features. Based on such conditions, the adversarial process of SCAL is trained to align conditional distributions to preserve the compactness of the target local structures. However, due to the non-differentiable property of the clustering procedure, it is hard to adapt the obtained local structures to the changing feature distribution in the training process. To greatly facilitate the learning process, we use a differentiable neural network as a surrogate for the clustering condition learner, leading to a differentiable end-to-end learning pipeline.

The main contributions of this work are summarized as follows.
1) We formulate unsupervised domain adaptation from the perspective of intrinsic local structure establishment by the clustering algorithm, which seeks to improve the class-structure-aware capability of domain adaptation.
2) We propose a structure-conditioned adversarial learning scheme (SCAL), which incorporates the clustering structure information into the adversarial learning process structure-aware conditions.
Besides, we use a differentiable neural network as a surrogate for a clustering learner to ensure the learning process to be end-to-end.
3) Experiments over several benchmark datasets demonstrate that the proposed scheme achieves significant performance gains in most cases.

\section{Related Work}
Recently, there has been many works~\cite{hsu2020progressive,hoffman2017cycada,tzeng2014deep,chen2020domain,yan2019weighted} on unsupervised domain adaptation (UDA) with deep models. The most relevant mainstream strategies of them to our research are adversarial methods and pseudo-labeling-aided methods, respectively. In the following, we discuss those relevant UDA methods in detail.


\subsection{Adversarial Domain Adaptation} 
In UDA, a typical line of research is based on adversarial methods. 
Inspired by the success of generative adversarial networks~\cite{goodfellow2014generative}, Ganin et al.~\cite{ganin2014unsupervised,ganin2016domain} introduce the adversarial methods to align features of different domains. They utilize a domain classifier (discriminator) to predict the domain labels, and a feature generator to generate an aligned distribution between domains to confuse the domain classifier. Therefore, the feature distribution discrepancy between different domains would be minimized. Following such a framework, Tzeng et al.~\cite{tzeng2017adversarial} propose the adversarial discriminative domain adaptation (ADDA) to utilize a shared domain classifier to align the outputs of two independent domain feature generators. SimNet~\cite{pinheiro2018unsupervised} introduces a similarity-based classifier and an adversarial loss to enhance the adaptation performance. Zhang et al.~\cite{zhang2019self} use multiple domain classifiers for collaborative adversarial learning. Without the classic domain classifier, Saito et al.~\cite{saito2018maximum} and followers~\cite{lee2019drop,saito2018adversarial} propose to utilize the adversarial process of two task classifiers to align domain distributions.

Besides aligning marginal distribution, several methods also align conditional or joint distributions. Based on DANN~\cite{ganin2016domain}, Long et al.~\cite{long2018conditional} consider conditioning the domain classifier on the category classifier and propose Conditional Domain Adversarial Network (CDAN). They aim to capture the cross-covariance between feature representations and classifier predictions. 
pseudo-labeled target sample Cicek and Soatto~\cite{cicek2019unsupervised} also aim to align the joint distribution over domain and label by a joint predictor and classifier's prediction. Deng et al.~\cite{deng2019cluster} and Xie et al.~\cite{xie2018learning} try to combine the class center alignment (based on the source classifier predictions) with the marginal distribution alignment to match the center-based marginal distributions between domains.

However, most of the adversarial UDA methods globally narrow down the domain shift across domains and cannot effectively preserve the intra-class compactness of the target domain. Although some researchers discover the phenomenon, they~\cite{xie2018learning} simply utilize the predictions of the source classifier to model the conditional distribution of the target domain features, which
would suffer the domain shift problem. Different from theirs, based on a structure-conditioned adversarial learning scheme, we can model the target domain structure by its intrinsic characteristics and then preserve its compactness.

\subsection{Pseudo-labeling-aided Domain Adaptation} 
Pseudo-labeling-aided methods have also gained great attention recently. In this kind of method, the researchers iteratively assign a pseudo-label to each unlabeled target data and use the pseudo-labeled target samples together with the source samples to learn an improved classification model. The obtained pseudo-labels are expected to be progressively more accurate when the model is optimized. For example, Long et al.~\cite{long2013transfer,zhang2017joint,wang2018visual} propose to obtain the pseudo-labels of target data by a classifier trained on the source samples. Sener et al.~\cite{sener2016learning} utilize KNN to obtain the pseudo-labels of target data. Pei et al.~\cite{pei2018multi} assign the conditional probability of each class to each target sample, which results in a soft label. 

Unfortunately, the performance of pseudo-labeling-aided methods relies heavily on the accuracy of the pseudo-labeling. Due to such a limitation, these methods may perform poorly when data distributions embody complex multimodal structures or need to rely on the complex noise reduction methods~\cite{kang2019contrastive,tang2020unsupervised,wang2020unsupervised} (e.g. sample selection). As a result, Chen et al.~\cite{chen2019progressive} introduce easy-to-hard sample selection method. Wang et al.~\cite{wang2020unsupervised} utilize a class-wise selection of two pseudo-labeling methods. Kang et al.~\cite{kang2019contrastive} introduce a Contrastive Adaptation Network (CAN), which only selects the categories with sufficient samples for training. 

\begin{figure*}[t]
	\centering
	\subfigure[Before Adaptation]{
		\label{fig:10cl:a}
		\begin{minipage}[ht]{0.30\textwidth}
			\includegraphics[width = 1\textwidth]{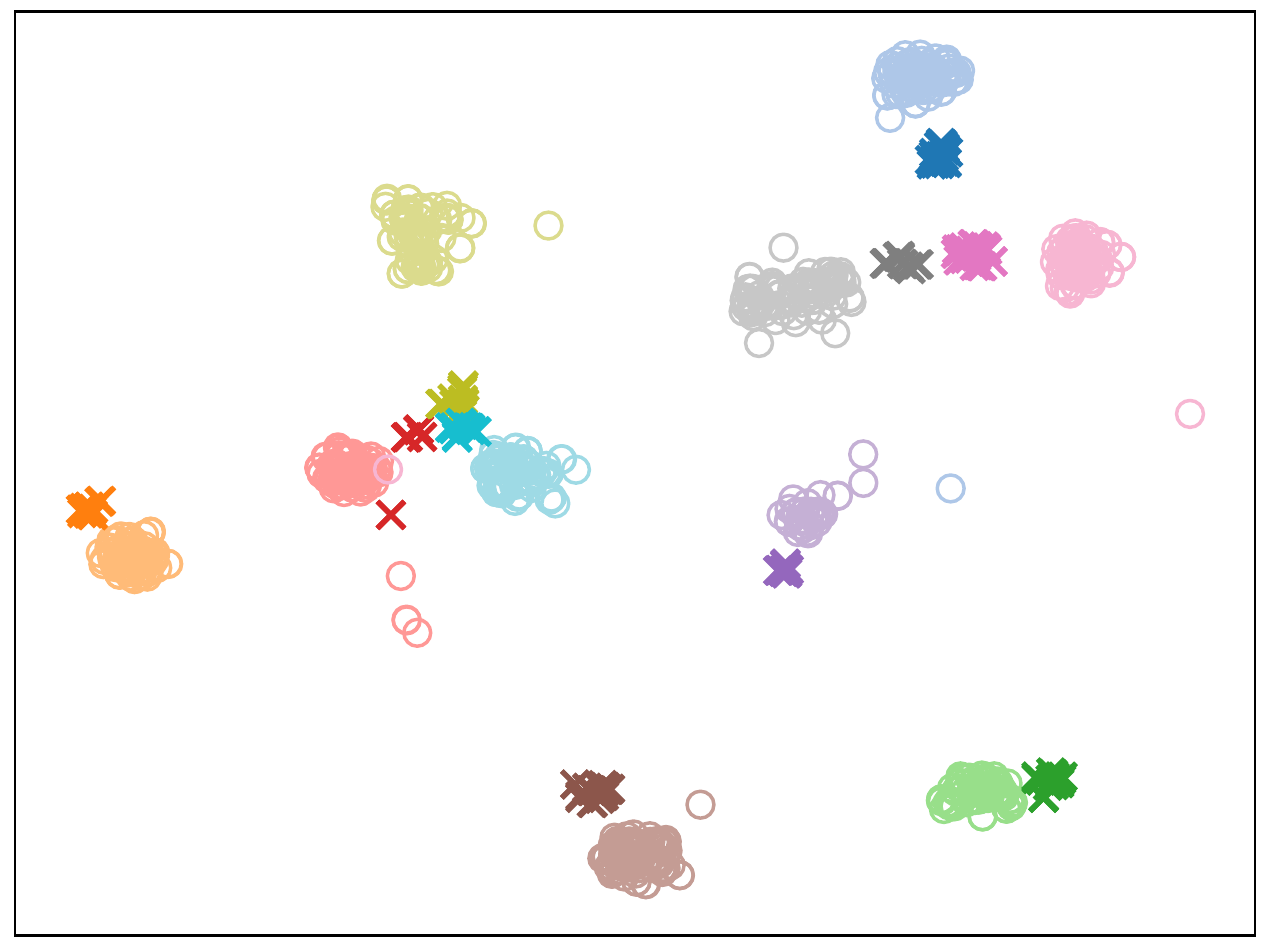}	
	\end{minipage}}
	\subfigure[Adversarial method]{
		\label{fig:10cl:b}
		\begin{minipage}[ht]{0.30\textwidth}
			\includegraphics[width = 1\textwidth]{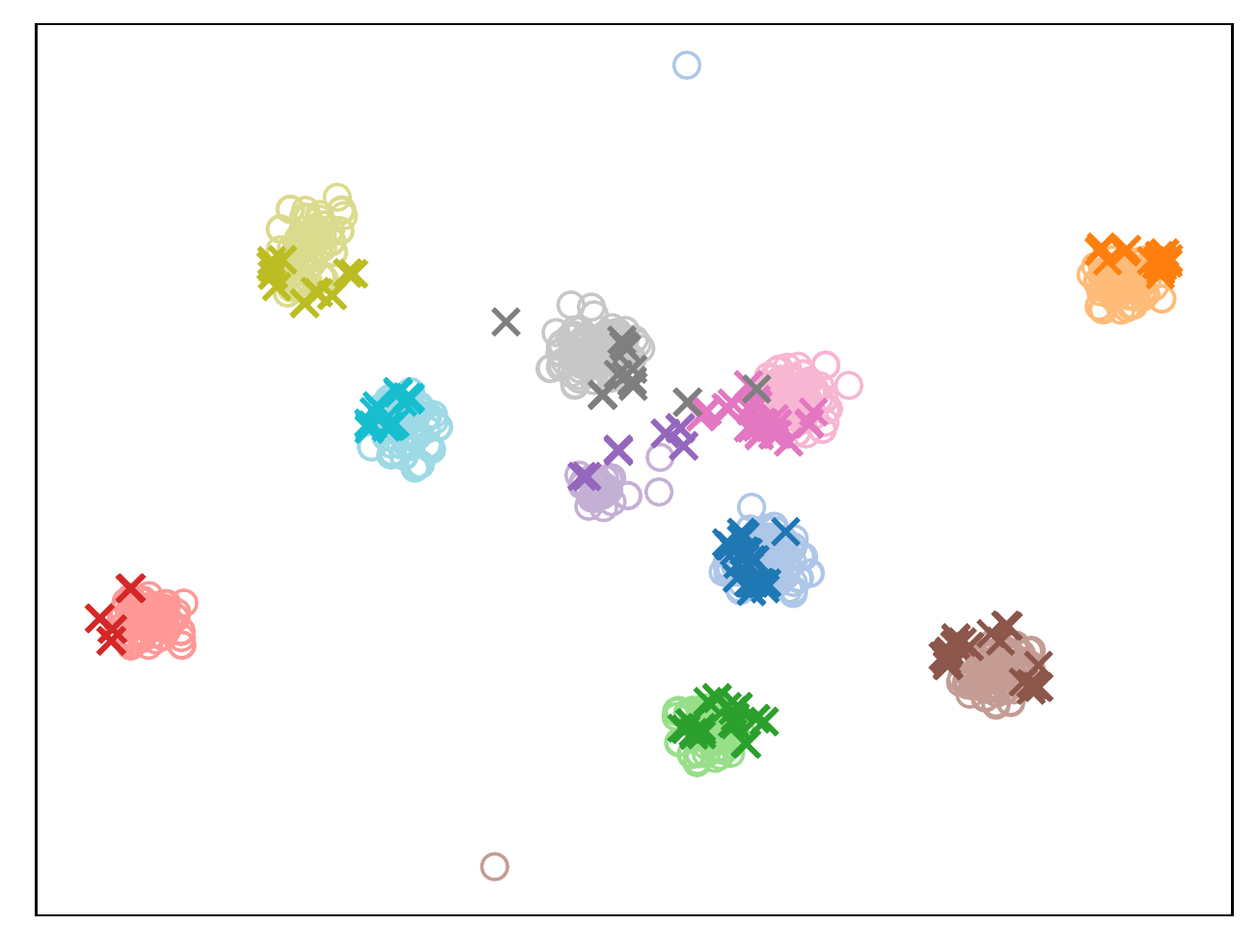}
		\end{minipage}	
	}
	\subfigure[Our Method]{
		\label{fig:10cl:c}
		\begin{minipage}[ht]{0.30\textwidth}
			\includegraphics[width = 1\textwidth]{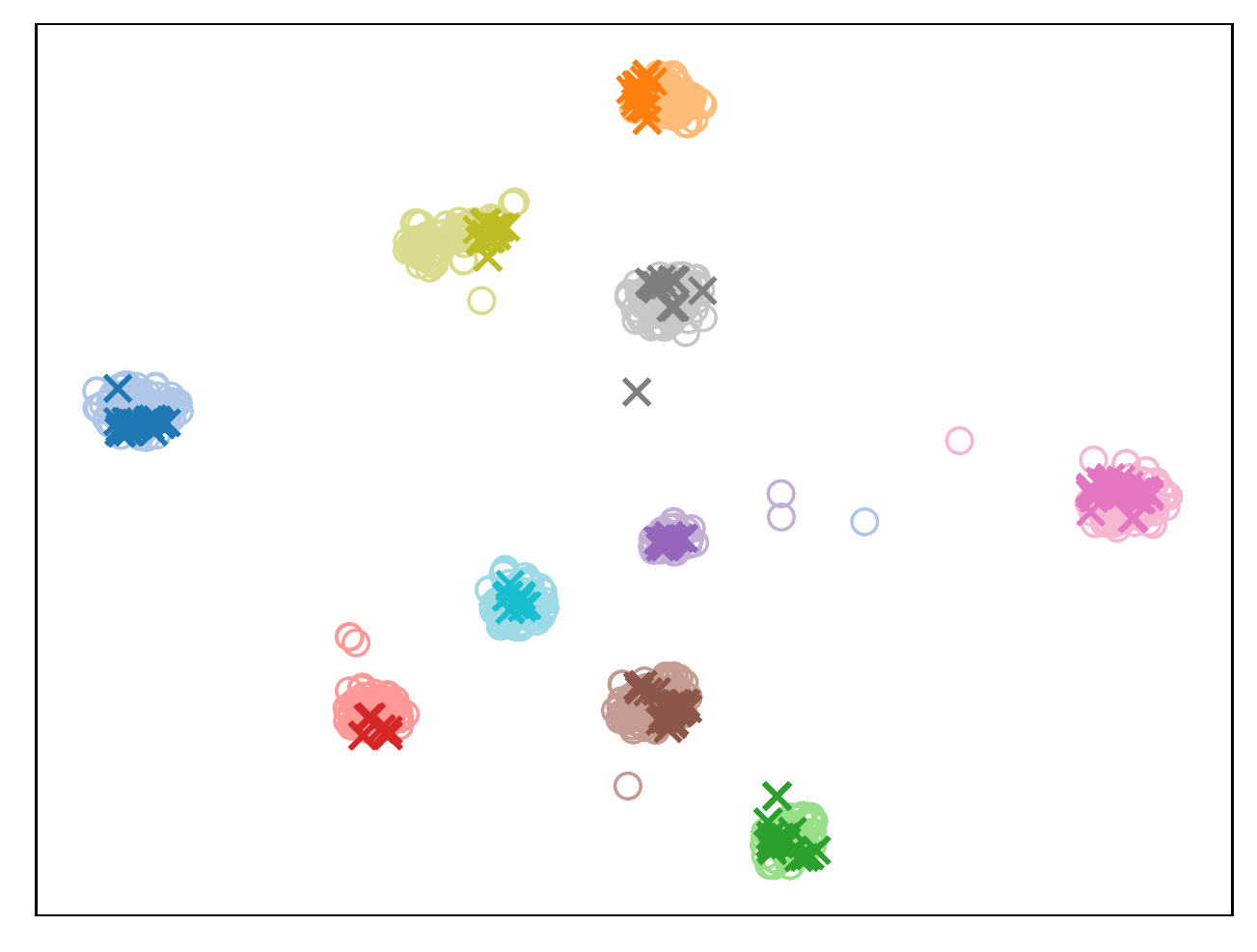}	
		\end{minipage}
	}
	\caption{Real feature visualization of different methods from the source domain \textbf{A} to the target domain \textbf{W} (i.e., \textbf{A}$\rightarrow$\textbf{W}) on Office-31~\cite{saenko2010adapting} with 10 classes ($\circ$:\textbf{A};$\times$:\textbf{W}). Here, the color of each point denotes its class, and the local structures of the features refer to the clusters of the features.}
	\label{fig:10cl}
\end{figure*}

In this paper, we are interested in employing the pseudo-labeling method to assist adversarial unsupervised domain adaptation. We propose a SCAL model to take the pseudo-labels of target domain features as conditions, thus preserving the dense clustered feature spaces in the adversarial process. Due to the statistics-based adversarial loss, our method is not sensitive to noise caused by clustering. We would discuss this in ~\cref{ea}.

\subsection{Metric Learning for Domain Adaptation} 
A growing number of researchers propose to reduce domain shift by metric learning. The general assumption of such methods is that the domain shift across domains can be measured by some distance metrics. After that, the domain shift reduction can be approximated as the optimization of the metric-based loss. For instance, some researchers~\cite{long2015learning,long2017deep,tzeng2014deep} utilize maximum mean discrepancy as a measure of the shift between the two domains. Sun et al.~\cite{sun2016return} and Li et al.~\cite{li2020gmfad} try to minimize the F-norm of covariance between different domains. Sun et al.~\cite{sun2016deep} introduce a method to match the mean and covariance across domains. Li et al.~\cite{li2020maximum} propose a maximum density divergence which aims to jointly minimizes the inter-domain divergence and maximizes the intra-domain density, while Luo et al.~\cite{luo2020unsupervised} adopt the affine Grassmann distance and the Log-Euclidean metric in unsupervised domain adaptation.

Apart from the above metrics, optimal transport~\cite{damodaran2018deepjdot} attracts much attention in recent years. Shen et al.~\cite{shen2018wasserstein} utilizes Wasserstein distance as the metric loss in promoting similarities between the features of different domains. Damodaran et al.~\cite{damodaran2018deepjdot} propose to transport the source samples to the target domain by an estimated mapping. Li et al.~\cite{li2020enhanced} introduce an enhanced optimal transport framework for domain adaptation. These optimal transport-based methods focus on global matching between source and target distributions, while we aim to model the local structures of target features by clustering.

\section{Proposed Method}
In this section, we illustrate our framework, called structure-conditioned adversarial learning (SCAL), in detail. First, we formulate the problem of SCAL in~\cref{pln}. Second, we elaborate on how to form the structure-conditioned adversarial loss of SCAL in~\cref{pd}. Third, we describe the two main components (i.e., local structure establishment and surrogate classifier approximation) of SCAL in~\cref{cpl}. Finally, we make a theoretical analysis of SCAL based on domain adaptation theory~\cite{ben2007analysis} in \cref{TI}.

\subsection{Problem Formulation}\label{pln}
Given a labeled source domain $\mathcal{D}^s$ and an unlabeled target domain $\mathcal{D}^t$, our goal is to train a network using $\mathcal{D}^s$ and $\mathcal{D}^t$ to make predictions on $\mathcal{D}^t$. Note that $\mathcal{D}^s$ and $\mathcal{D}^t$ have the same $K$-label space $\{1,2,\dots,K\}$, and the data of them are drawn from different distributions, which leads to a domain shift~\cite{ben2007analysis,ben2010theory,torralba2011unbiased}.

Our motivation is to reduce the domain shift and preserve the local structures of the target domain simultaneously. For better illustration, we take \cref{fig:10cl} as an example. As shown in \cref{fig:10cl}, on task \textbf{A}$\rightarrow$\textbf{W}, it is notable that the variance of the target local structures increases after the adversarial domain adaptation~\cite{ganin2016domain} (the $\times$ of~\cref{fig:10cl:a} vs. the $\times$ of~\cref{fig:10cl:b}), which makes the classifier difficult to perform the classification of target domain data. As a result, we propose SCAL, which aims to preserve the target local structures, i.e., the clusters of the target features, to improve the discriminability of the adapted target features. 

Concretely, SCAL consists of a feature extractor $G$, a source classifier $F$, a surrogate classifier $F_S$, and a domain classifier $D$. And the training process of SCAL is a minimax game:
\begin{gather}
\begin{aligned}
\mathop{\min}\limits_{G, F}\mathop{\max}\limits_{D}\mathcal{L}_{cls} - \lambda\mathcal{L}_{scal},\label{overall}
\end{aligned}
\end{gather}
where $\mathcal{L}_{cls}$ denotes the classification loss and $\mathcal{L}_{scal}$ denotes the structured-conditioned adversarial loss. $\mathcal{L}_{cls}$ is the cross-entropy loss $\mathcal{L}_{ce}$ in source domain with ground-truth labels:
\begin{gather}
\mathcal{L}_{cls}=\mathbb{E}_{(x,y)\sim\mathcal{D}^s}[\mathcal{L}_{ce}(F(G(x)),y)]. \label{s-cls}
\end{gather}


\begin{figure*}[t]
	\centering
	\includegraphics[width=\textwidth]{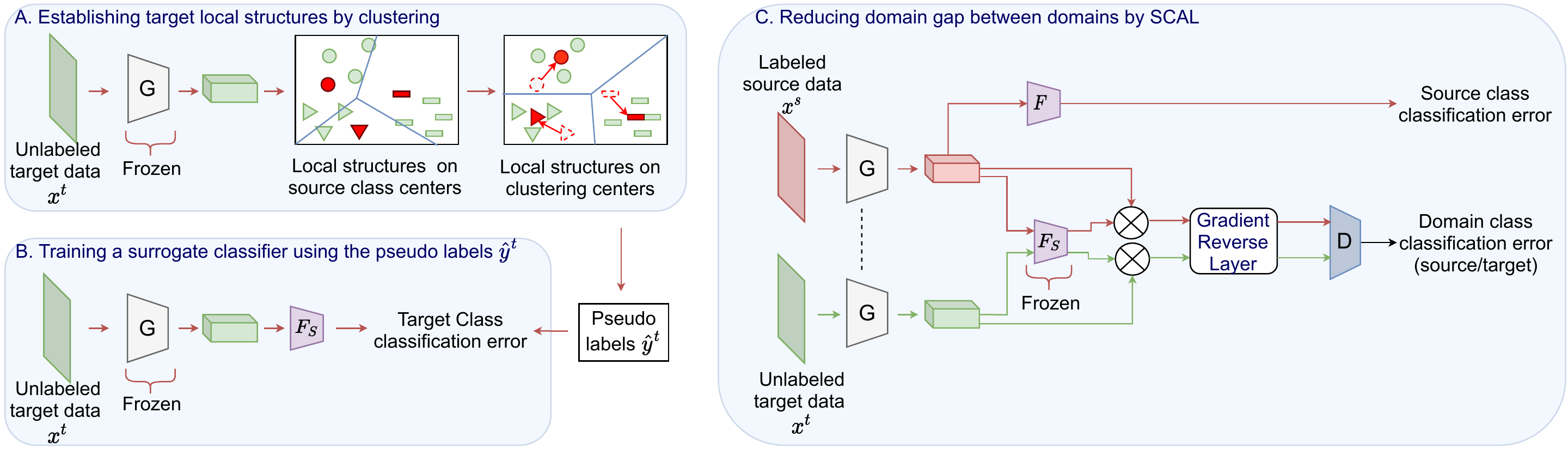}
	\caption{The framework of structure-conditioned adversarial learning (SCAL). Here, Gradient Reverse Layer~\cite{ganin2014unsupervised,ganin2016domain} is used to make the gradients of its subsequent module $D$ opposite to the gradients of its previous module $G$, therefore achieving adversarial training in one backpropagation process. Before each epoch, we establish the target local structures by clustering in step A (\cref{cpl}). For each batch of one epoch, we adopt two steps. We first employ a surrogate classifier to mimic the non-differentiable clustering procedure in step B. Then the predictions of the surrogate classifier are used as structure-aware conditions for the adversarial learning process in step C.}
	\label{fig:biggraph-full}
	\vspace{-5pt}
\end{figure*}

\subsection{Structure-Conditioned Adversarial Loss}\label{pd}
The idea behind SCAL is to preserve the local structures between domains in the global alignment process of adversarial learning. With this motivation, we propose to extract the structure-conditioned features of target domain data and incorporate them into the adversarial loss:
\begin{gather}
\begin{aligned}
\mathcal{L}_{scal}=&-\mathbb{E}_{x\sim\mathcal{D}^s}[\log D(S(x))]\\
&-\mathbb{E}_{x\sim\mathcal{D}^t}[\log(1-D(S(x)))],
\label{scd-cls}
\end{aligned}
\end{gather}
where $S$ denotes the extractor of the structure-conditioned features. Note that~\cref{scd-cls} defines the adversarial loss on the structure-conditioned features. By using the conditioning strategy, the structure-aware conditions have been successfully applied to the feature distribution. As a result, the adversarial process could potentially align the features that have similar structure-aware conditions, thereby achieving the goal of preserving the compactness of the target local structures.

To extract the structure-conditioned feature $S(x)$, we need a function that outputs the pseudo-label conditions $\hat{y}$, which can represent the local structures of target features. As shown in \cref{fig:biggraph-full}, we achieve it by two steps: local structure establishment and surrogate classifier approximation. Indeed, by doing these two steps, we are making a differentiable approximation, i.e, a surrogate classifier $F_S$ to the non-differentiable local structures for the target domain. We describe the two steps in the next subsection.

\begin{algorithm}[!b]
	\caption{Class-aware Memory-base Domain adaptation}
	\label{procedure}
	\begin{algorithmic}[1]
		\REQUIRE
		The source domain dataset $\mathcal{D}^s$, the target domain dataset $\mathcal{D}^t$;
		\ENSURE
		The learned parameters of $G$, $F$, and $F_S$;
		\STATE Initialize a feature extractor $G$ by an ImageNet pre-trained ResNet\;
		\STATE Randomly initialize a source classifier $F$ and a surrogate classifier $F_S$\; 
		\FOR{epochs $1, 2, 3, \dots, \mathrm{MAX\_EPOCH}$}
		\STATE Compute source class centers over the source features\;
		\STATE Initialize target cluster centers by the source class centers\;
		\STATE Obtain the pseudo-labels of target samples by the spherical K-means (\cref{cpl})\;
		\FOR{iterations $1, 2, 3, \dots, \mathrm{MAX\_ITER}$}
		\STATE Sample a source data batch $B_s$ and a target data batch $B_t$ from $\mathcal{D}^s$ and $\mathcal{D}^t$\;
		\STATE Update $F_S$ by~\cref{surrogate} on $B_t$\;
		\STATE Update $G$ and $F$ by~\cref{overall} on $B_s$ and $B_t$\;		
		\ENDFOR
		\ENDFOR
	\end{algorithmic}
\end{algorithm}

\subsection{Local Structures for Target Domain}\label{cpl}
Based on~\cref{scd-cls}, we need to establish the local structures, i,e,, the feature clusters, for the unlabeled target domain data. Mathematically, searching the feature clusters could be cast as the following optimization problem:
\begin{align}
\mathop{\min}\limits_{\{C^t_k\}_{k=1}^K}\sum_{k=1}^{K}\sum_{x\in C^t_k}\mathcal{L}_{dist}(G(x),\mu_k^t),\label{clustering_form}
\end{align}
where $C^t_k$ denotes the learned target cluster for class $k$, $\mu_k^t$ denotes the center of $C^t_k$, and the distance metric satisfies:
\begin{align}
\mathcal{L}_{dist}(G(x),\mu_k^t)=\frac{1}{2}(1-\frac{<G(x),\mu_k^t>}{\left\|G(x)\right\|\left\|\mu_k^t\right\|}).
\end{align}

A common approach for \cref{clustering_form} is spherical K-means~\cite{dhillon2001concept,zhong2005efficient}. Inspired by it, we adopt a source class center initialized clustering algorithm, as shown in~\cref{fig:biggraph-full}, especially for the UDA scenarios (considering the inherent relation between domains). With such a clustering algorithm, we can obtain the target clusters as well as the clustering labels for all target samples.

Note that the above local structures establishment process is non-differentiable. Because of it, it is hard to adapt the obtained local structures to the changing feature distribution in the training process. The fixed clustering result is harmful for the final performance, as shown in \cref{tab:aba}. 

To deal with this issue, we provide a differentiable approximation, i.e., a differentiable surrogate classifier $F_S$, to approximate the obtained local structures. The differentiable surrogate classifier $F_S$ is updated at each batch by minimizing the cross-entropy loss $\mathcal{L}_{ce}$ on the obtained pseudo-labeled target sample set $\hat{\mathcal{D}}^t$:
\begin{equation}
\mathop{\min}\limits_{F_S}\mathbb{E}_{(x,\hat{y})\sim\hat{\mathcal{D}}^t}[\mathcal{L}_{ce}(F_S(G(x)),\hat{y})].
\label{surrogate}
\end{equation}


Then the structure-conditioned features $S(x)$ for each instance $x\in\mathcal{D}^s\cup\mathcal{D}^t$ can be obtained by taking the predictions of the surrogate classifier as conditions:
\begin{gather}
\begin{aligned}
S(x)=G(x)\otimes F_S(G(x)).
\end{aligned}
\end{gather}
where $\otimes$ denotes outer product. Therefore, the overall structured-conditioned loss $\mathcal{L}_{scal}$ of SCAL can be formulated as:
\begin{gather}
\begin{aligned}
\mathcal{L}_{scal}=&-\mathbb{E}_{x\sim\mathcal{D}^s}[\log D(G(x)\otimes F_S(G(x)))]\\
&-\mathbb{E}_{x\sim\mathcal{D}^t}[\log(1-D(G(x)\otimes F_S(G(x))))].
\label{scd-cls-2}
\end{aligned}
\end{gather}


\begin{figure}[t]
	\centering
	\includegraphics[width=1.0\columnwidth]{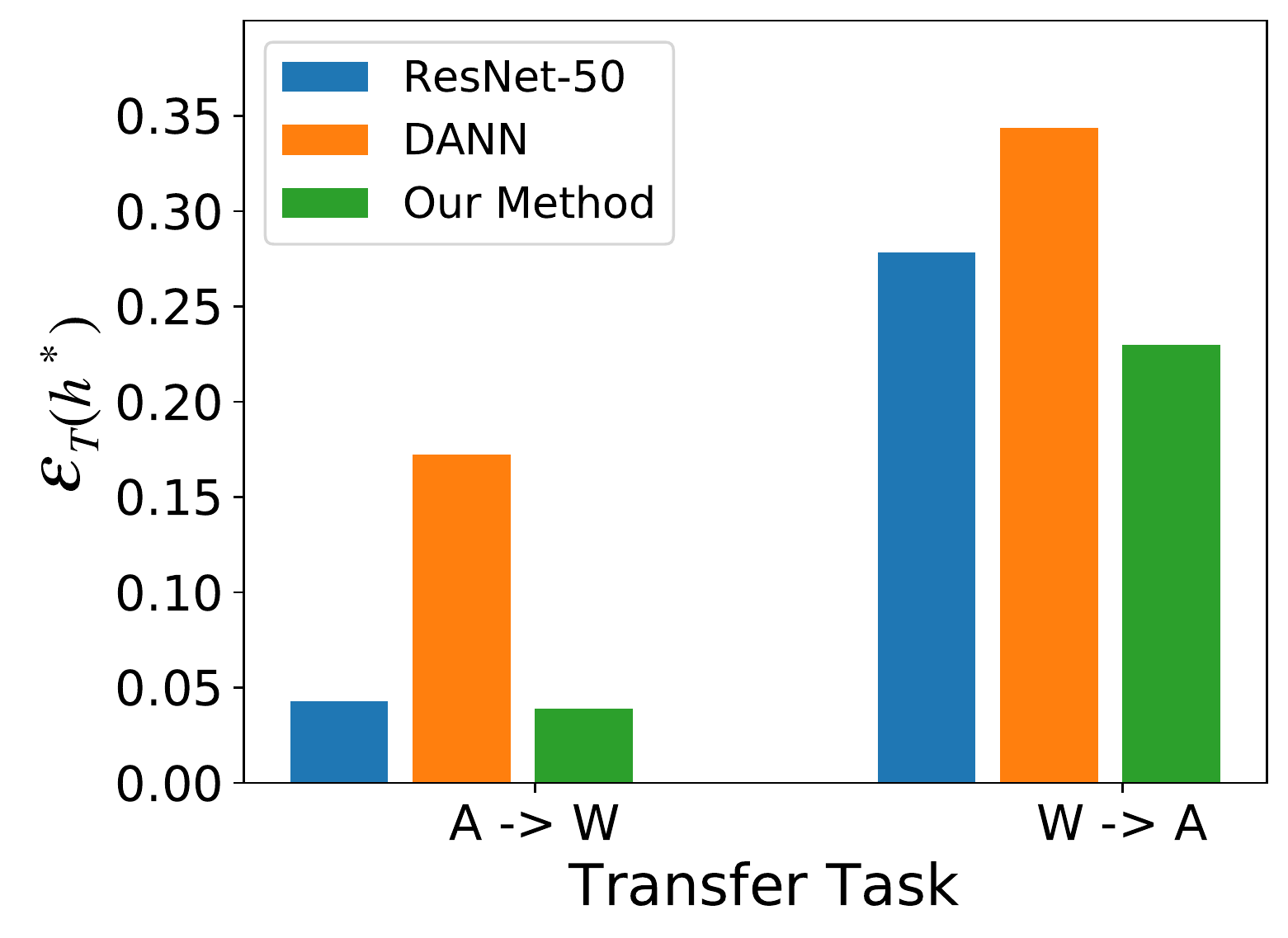}
	\caption{The expected error $\mathcal{E}_T(h^*)$ on the target features of the ideal joint hypothesis $h^*$ for different methods.}
	\label{fig:dis}
\end{figure}

\begin{figure*}[!t]
	\centering
	\subfigure[Office-31]{
		\label{fig:office-31:a}
		\begin{minipage}[ht]{0.28\textwidth}
			\includegraphics[width = 1\columnwidth]{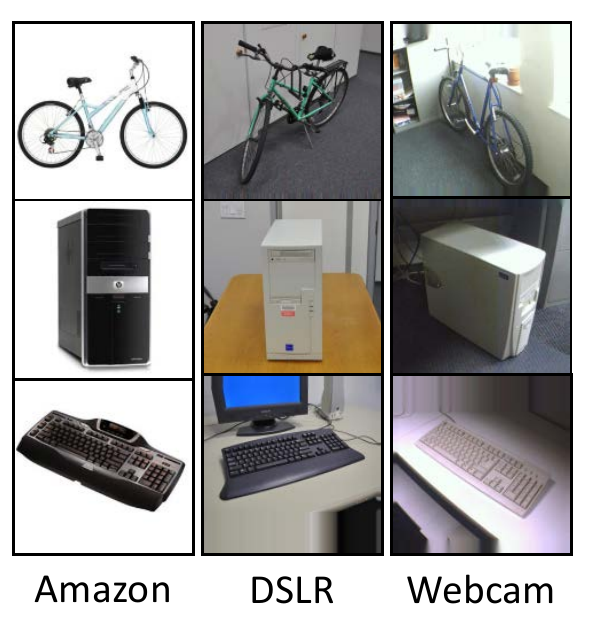}
		\end{minipage}	
	}	\subfigure[Office-Home]{
		\label{fig:office-home:b}
		\begin{minipage}[ht]{0.38\textwidth}
			\includegraphics[width = 1\columnwidth]{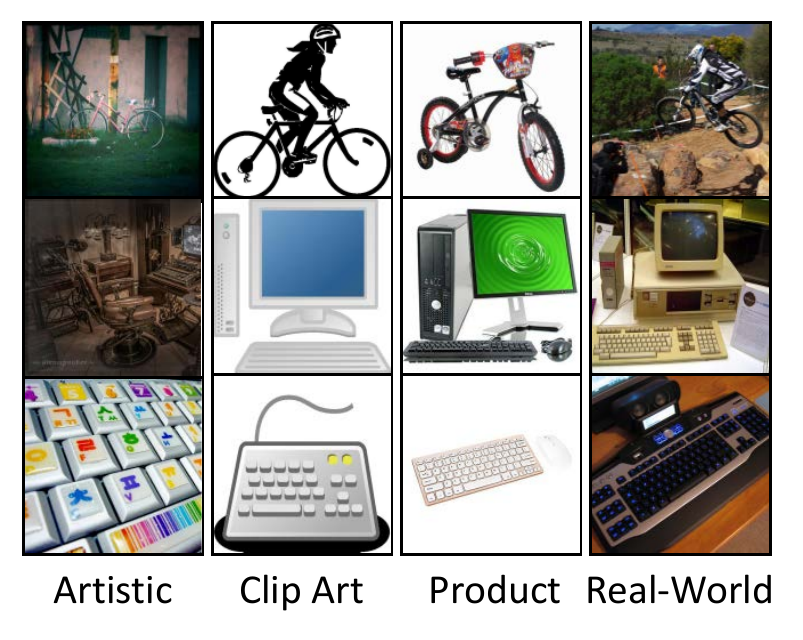}
		\end{minipage}	
	}
	\subfigure[VisDA-2017]{
		\label{fig:visda:c}
		\begin{minipage}[ht]{0.28\textwidth}
			\includegraphics[width = 1\columnwidth]{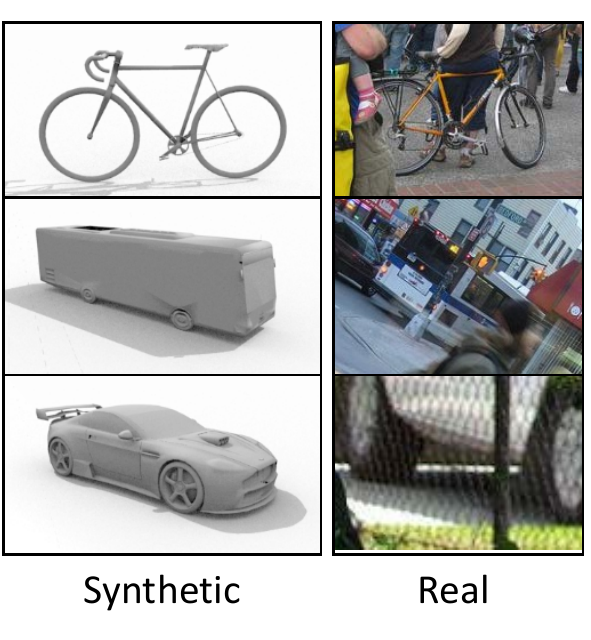}	
		\end{minipage}
	}
	\caption{Exemplary images from different datasets. (a) Office31: It has three distinct domains, including Amazon, DSLR, and Webcam. (b) Office-Home: It has four dissimilar domains, including Artistic, Clip Art, Product, and Real-World. (c) VisDa-2017: It has two domains, including Synthetic and Real.}
	\label{fig:dataset}
\end{figure*}

For a clear description of our implementation procedure, we provide the algorithm flow of SCAL in~\cref{procedure}. Our method mainly consists of three steps: local structure establishment, surrogate classifier approximation, and structure-conditioned adversarial learning. For each epoch, we firstly establish the local structure by a source center initialized clustering algorithm (lines 4-6). And for each batch of one epoch, we use the surrogate classifier to approximate the local structure $F_S$ (line 9) and then integrate it into an adversarial process (line 10) as structure conditions.

\subsection{Theoretical Analysis}\label{TI}
The theoretical analysis of SCAL is based on the domain adaptation theory~\cite{ben2007analysis}. The theory reveals that the expected error $\mathcal{E}_T(h)$ on target samples of any classifier $h$ (drawn from a hypothesis set $\mathcal{H}$ with optimal classifier $h^*$) has the following upper-bound w.r.t. the source error $\mathcal{E}_S(\cdot)$:
\begin{equation}
\begin{aligned}
\forall h \in \mathcal{H}, \mathcal{E}_T(h)\le \mathcal{E}_S(h)+\frac{1}{2}d_{\mathcal{H}}
+\mathcal{E}_S(h^*) + \mathcal{E}_T(h^*).
\label{error_bound}
\end{aligned} 
\end{equation}

In fact, previous adversarial methods mainly focus on the domain discrepancy
$d_{\mathcal{H}}$ (Note that $\mathcal{E}_S(h)$ and $\mathcal{E}_S(h^*)$ can be controlled by the classification loss on labeled source data) and usually ignore the discriminability of the target features $\mathcal{E}_T(h^*)$. In contrast, 
we aim to pursue the balance between $d_{\mathcal{H}}$ and $\mathcal{E}_T(h^*)$ for UDA, which is detailed in the following sections.

\subsubsection{Analysis for Target Feauture Discriminability}\label{app:dis}

To investigate target feature discriminability $\mathcal{E}_T(h^*)$ in-depth, we use a well-trained multilayer perceptron (MLP) classifier to measure the $\mathcal{E}_T(h^*)$ for different methods. For convenience, we take the adversarial domain adaptation method, i.e., DANN, for example. In domain adaptation theory~[2], the ideal joint hypothesis $h^*$ is defined as:
\begin{equation}
\begin{aligned}
h^*= \mathop{\min}_{h\in\mathcal{H}}\mathcal{E}_S(h)+\mathcal{E}_T(h)
\end{aligned}
\end{equation}
which is the optimal classifier $h^*$ for the source dataset and the target dataset. To obtain the $h^*$, we train a MLP with the fixed representations outputted by the feature extractor of DANN over all data with the labels from both domains. Note that the target labels are only used for this analysis. Then the $\mathcal{E}_T(h^*)$ of DANN is expected to be the average prediction error of the MLP, i.e., $h^*$, on the fixed representations of the target domain data. The process of obtaining $\mathcal{E}_T(h^*)$ in other methods is the same as the above process with results shown in~\cref{fig:dis}. 

From ~\cref{fig:dis}, we observe that the $\mathcal{E}_T(h^*)$ is reduced after adversarial domain adaptation by DANN~[11]. Obviously, a higher error rate implies weaker discriminability, which means the classifier is difficult to perform the classification of target domain data. With the cluster conditions, our method can preserve the discriminability of original feature distribution as shown in~\cref{fig:dis}, therefore lowering the upper-bound of the expected error $\mathcal{E}_T(h)$. 

\subsubsection{Analysis for the Domain Discrepancy}\label{app:theorem}
Based on domain theory~[2], the general domain discrepancy $d_{\mathcal{H}}$ is measured over the feature space:
\begin{equation}
\begin{aligned}
d_{\mathcal{H}}=&2\mathop{\text{sup}}\limits_{h\in\mathcal{H}}\big|\mathbb{E}_{x\sim\mathcal{D}^s(G)}[\text{I}(h(x)=1)]\\
&-\mathbb{E}_{x\sim\mathcal{D}^t(G)}[\text{I}(h(x)=1)]\big|,
\end{aligned}
\end{equation}
where $\mathcal{D}^s(G)\!=\!\{G(x)|x\!\in\!\mathcal{D}^s\}$ and $\mathcal{D}^t(G)\!=\!\{G(x)|x\!\in\!\mathcal{D}^t\}$ denote the set of the source and target features respectively. Note a hypothesis is a function $h\!:\!\mathcal{X}\!\rightarrow\!\{0,1\}$, and $\mathcal{H}$ is the hypothesis space. We then define the $\mathcal{H}_D$-distance over the structure-conditioned feature space as:
\begin{equation}
\begin{aligned}
d_{\mathcal{H}_D}=&2\mathop{\text{sup}}\limits_{D\in\mathcal{H}_D}\big|\mathbb{E}_{x\sim\mathcal{D}^s(S)}[\text{I}(D(x) = 1)]\\
&-\mathbb{E}_{x\sim\mathcal{D}^t(S)}[\text{I}(D(x) = 1)]\big|,\label{dh}
\end{aligned}
\end{equation}
where
\begin{equation}
\begin{aligned}
S(x)=G(x)\otimes F_S(G(x)).
\end{aligned}
\end{equation}
In the above equation, $\mathcal{D}^s(S)\!=\!\{S(x)|x\!\in\!\mathcal{D}^s\}$ and $\mathcal{D}^t(S)\!=\!\{S(x)|x\!\in\!\mathcal{D}^t\}$ denote the structure-conditioned feature set of the source and target domains, respectively.

\textbf{Theorem 1.} Let $D$ be a hypothesis in a symmetric hypothesis space $\mathcal{H}_D$ (If $D\in\mathcal{H}_D$, the inverse hypothesis $1-D$ is also in $\mathcal{H}_D$), then following inequality is satisfied:
\begin{equation}
d_{\mathcal{H}} \le d_{\mathcal{H}_D}.
\end{equation}

\emph{Proof} Formally, for a sample $x$, given the output of the feature generator $G(x)=(g_1,g_2,\dots,g_N)^T$, the surrogate classifier prediction $F_S(G(x))=(f_1,f_2,\dots,f_M)^T$, and an all-ones vector $J_M=(\underbrace{1,1,\dots,1}_M)^T$, we have
\begin{small}
	\begin{equation}
	\begin{aligned}
	S(x)J_M&=(G(x)\otimes F_S(G(x)))J_M=G(x)\sum_{i=1}^{M}f_i.
	\label{transfer}
	\end{aligned} 
	\end{equation}
\end{small}

For any hypothesis $h\in\mathcal{H}$, we can construct a corresponding hypothesis $D_{h}(x)=h(\frac{xJ_M}{\sum_{i=1}^{M}f_i})$. Assuming the family of domain classifier $\mathcal{H}_D$ is rich enough to contain $D_{h}(x)$, i.e., $D_{h}(x)\in\mathcal{H}_D$. Such an assumption is realistic as we can use a complex domain classifier $D$, i.e., the multilayer perceptrons, to fit any functions. Then from \cref{transfer}, we have

\begin{small}
	\begin{equation}
	\begin{aligned}
	\label{d2}
	D_h(S(x))= h(\frac{S(x)J_M}{\sum_{i=1}^{M}f_i})=h(G(x)).
	\end{aligned} 
	\end{equation}
\end{small}

From \cref{dh,d2}, the following inequality can be derived:
\begin{small}
	\begin{equation}
	\begin{aligned}
	\label{d3}
	d_{\mathcal{H}_D}\!&=\! 2\mathop{\text{sup}}_{D\in\mathcal{H}_D}\big|\mathbb{E}_{x\sim\mathcal{D}^s(S)}[\text{I}(D(x)\!=\!1)]
	\!-\!\mathbb{E}_{x\sim\mathcal{D}^t(S)}[\text{I}(D(x)\!=\!1)]\big|\\
	&\ge\!2\big(\mathbb{E}_{x\sim\mathcal{D}^s(S)}[\text{I}(D_h(x)\!=\! 1)]\!-\!\mathbb{E}_{x\sim\mathcal{D}^t(S)}[\text{I}(D_h(x)\!=\!1)]\big)\\
	&=\!2\big(\mathbb{E}_{x\sim\mathcal{D}^s}[\text{I}(D_h(S(x))\!=\! 1)]\!-\!\mathbb{E}_{x\sim\mathcal{D}^t}[\text{I}(D_h(S(x))\!=\!1)]\big).\\
	&\!=\!2\big(\mathbb{E}_{x\sim\mathcal{D}^s}[\text{I}(h(G(x))\!=\! 1)]\!-\!\mathbb{E}_{x\sim\mathcal{D}^t}[\text{I}(h(G(x))\!=\!1)]\big) \\ 
	&\!=\!2\big(\mathbb{E}_{x\sim\mathcal{D}^s(G)}[\text{I}(h(x)\!=\! 1)]\!-\!\mathbb{E}_{x\sim\mathcal{D}^t(G)}[\text{I}(h(x)\!=\!1)]\big). 
	\end{aligned} 
	\end{equation}
\end{small}

Considering \cref{d2} and the symmetry of $\mathcal{H}_D$, i.e., $1-D_h(x)\in\mathcal{H}_D$, we have
\begin{small}
	\begin{equation}
	\begin{aligned}
	\label{d4}
	d_{\mathcal{H}_D}\!
	&\!\ge-2\big(\mathbb{E}_{x\sim\mathcal{D}^s(G)}[\text{I}(h(x)\!=\! 1)]-\mathbb{E}_{x\sim\mathcal{D}^t(G)}[\text{I}(h(x)\!=\!1)]\big).
	\end{aligned} 
	\end{equation}
\end{small}

Since for each $h\in\mathcal{H}$, \cref{d3,d4} are satisfied, the distribution discrepancy $d_{\mathcal{H}_D}$ with a symmetric hypothesis space $\mathcal{H}_D$ upper-bounds the general distribution discrepancy $d_{\mathcal{H}}$. 

It is worth noting that the objective of the minimax game we solve in our method is to minimize $d_{\mathcal{H}_D}$. Concretely, the domain classifier $D$ is trained to distinguish source and target sample (obtain $d_{\mathcal{H}_D}$) and the extractor $G$ tries to minimize it. As a result, the bound, i.e., $d_{\mathcal{H}_D}$, of the domain discrepancy $d_{\mathcal{H}}$ is expected to be lowered in the training process of SCAL.

\section{Experiments}\label{exp}
\begin{table*}
	\caption{Accuracy (\%) for all the twelve transfer tasks of Office-Home based on ResNet-50. For simplicity, Ar$\rightarrow$Cl is denoted by ArCl, and the names of other transfer tasks are simplified similarly.}
	\label{tab:office-Home}
	\small
	\begin{tabu} to 0.99\textwidth{X[5, l] X[3c] X[c] X[c] X[c] X[c] X[c] X[c] X[c] X[c] X[c] X[c] X[c] X[c] X[c]}
		\toprule
		Method&Year&ArCl&ArPr&ArRw&ClAr&ClPr&ClRw&PrAr&PrCl&PrRw&RwAr&RwCl&RwPr&Avg\\
		\midrule
		ResNet-50~\cite{he2016deep}&CVPR'2016&34.9&50.0&58.0&37.4&41.9&46.2&38.5&31.2&60.4&53.9&41.2&59.9&46.1\\
		DANN~\cite{ganin2016domain}&JMLR'2016&45.6&59.3&70.1&47.0&58.5&60.9&46.1&43.7&68.5&63.2&51.8&76.8&57.6\\
		CDAN+E~\cite{long2018conditional}&NIPS'2018&50.7&70.6&76.0&57.6&70.0&70.0&57.4&50.9&77.3&70.9&56.7&81.6&65.8\\
		SymNets~\cite{zhang2019domain}&CVPR'2019&47.7&72.9&78.5&64.2&71.3&74.2&64.2&48.8&79.5&74.5&52.6&82.7&67.6\\
		BSP+CDAN~\cite{chen2019transferability}&ICML'2019&52.0&68.6&76.1&58.0&70.3&70.2&58.6&50.2&77.6&72.2&59.3&81.9&66.3\\
		SPL~\cite{wang2020unsupervised}&AAAI'2020&54.5&\textbf{77.8}&\textbf{81.9}&65.1&\textbf{78.0}&\textbf{81.1}&66.0&53.1&\textbf{82.8}&69.9&55.3&\textbf{86.0}&71.0\\
		DMP~\cite{luo2020unsupervised}&PAMI'2020&52.3&73.0&77.3&64.3&72.0&71.8&63.6&52.7&78.5&72.0&57.7&81.6&68.1\\
		ATM~\cite{li2020maximum}&PAMI'2020&52.4&72.6&78.0&61.1&72.0&72.6&59.5&52.0&79.1&73.3&58.9&83.4&67.9\\
		AADA+CCN~\cite{yangmind}&ECCV'2020&54.0&71.3&77.5&60.8&70.8&71.2&59.1&51.8&76.9&71.0&57.4&81.8&67.0\\
		GVB-GD~\cite{cui2020gradually}&CVPR'2020&\underline{57.0}&74.7&79.8&64.6&74.1&74.6&65.2&\underline{55.1}&81.0&74.6&\underline{59.7}&84.3&70.4\\
		RSDA-MSTN~\cite{gu2020spherical}&CVPR'2020&53.2&\underline{77.7}&\underline{81.3}&66.4&74.0&76.5&\underline{67.9}&53.0&\underline{82.0}&\underline{75.8}&57.8&\underline{85.4}&70.9\\
		SRDC~\cite{tang2020unsupervised}&CVPR'2020&52.3&76.3&81.0&\textbf{69.5}&76.2&78.0&\textbf{68.7}&53.8&81.7&\textbf{76.3}&57.1&85.0&71.3\\
		\midrule
		SCAL&-&55.3&72.7&78.7&63.1&71.7&73.5&61.4&51.6&79.9&72.5&57.8&81.0&68.3\\
		SCAL+SPL&-&\textbf{57.3}&77.5&80.7&\underline{68.8}&\underline{77.9}&\underline{79.3}&65.2&\textbf{55.9}&81.7&75.0&\textbf{61.0}&83.9&\textbf{72.0}\\
		\bottomrule
	\end{tabu}
\end{table*}

\subsection{Datasets}

\textbf{Office-Home}~\cite{venkateswara2017deep} is a challenging dataset, which contains around 15,500 images from 65 categories. It consists of images from four dissimilar domains: \emph{Artistic} (\textbf{Ar}), \emph{Clip Art} (\textbf{Cl}), \emph{Product} (\textbf{Pr}), and \emph{Real-World} (\textbf{Rw}). We evaluate our methods on all the twelve transfer tasks.

\textbf{Office-31}~\cite{saenko2010adapting} is the most widely used benchmark for unsupervised domain adaptation. It consists of $4110$ images belonging to $31$ categories collected from three distinct domains: \emph{Amazon} (\textbf{A}), \emph{Webcam} (\textbf{W}), \emph{DSLR} (\textbf{D}). In particular, the dataset is imbalanced across domains, with 2,817 images in \textbf{A}, 795 images in \textbf{W}, and 498 images in \textbf{D}. We evaluate our methods on all the six transfer tasks.

\textbf{VisDA-2017}~\cite{visda2017} is a large simulation-to-real dataset, which consists of over 280K images across 12 classes. It contains two very distinct domains: \textbf{Synthetic}, synthetic 2D renderings of 3D models generated from different angles and with different lighting conditions; \textbf{Real}, a photo-realistic or real-image domain. We evaluate our methods on \textbf{Synthetic}$\rightarrow$\textbf{Real} transfer task.

\subsection{Setup}

\textbf{Network Architectures.}
Similar to other unsupervised domain adaptation methods~\cite{ganin2016domain,long2018conditional}, we employ ResNet~\cite{he2016deep} (including ResNet-50 and ResNet-101) as our backbone network. Specifically, we utilize the feature extractor part of ResNet (before average pooling layer) as our feature extractor and one fully-connected layer as our source classifier. As for the domain classifier or surrogate classifier, we also adopt one fully-connected layer.

\textbf{Evaluation Protocol.}
We follow the standard evaluation protocol for the unsupervised domain adaptation~\cite{ganin2016domain,long2018conditional}. We use all labeled source domain samples and all unlabeled target domain samples during the training procedure, and choose the output of the surrogate classifier as our final prediction. We compare the mean classification accuracy based on three random experiments. 

\textbf{Training Details.} For all benchmarks, we implement our methods based on PyTorch and finetune our network with the feature extractor part of ResNet pre-trained on ImageNet~\cite{russakovsky2015imagenet}. And we train the network through back-propagation~\cite{lecun1990handwritten}, where the source classifier and the surrogate classifier are trained from scratch with a learning rate ten times that of the feature extractor part. For hyper-parameters of \emph{Office-31} and \emph{Office-Home}, we adopt mini-batch SGD with epoch number of 100, momentum of 0.9, $\lambda$ of 1, and learning rate decay strategy implemented in DANN~\cite{ganin2016domain}: the learning rate is updated by $\eta_p=\eta_0(1 + \alpha p)^{-\beta}$, where $p$ is the training process linearly changing from $0$ to $1$, and $\eta_0 = 0.001$, $\alpha=10$, $\beta=0.75$. As for \emph{VisDA-2017}, the difference is that we use a new learning rate decay strategy with $\eta_0 = 0.001$, $\alpha=5$, $\beta=2.25$.

\begin{table*}
	\caption{Accuracy (\%) for all the six transfer tasks of Office-31 based on ResNet-50. (*) the implementation is based on the authors' code.}
	\label{tab:office-31}
	\small
	\begin{tabu} to 0.99\textwidth{X[4] X[2c] X[2c] X[2c] X[2c] X[2c] X[2c] X[2c] X[c]}
		\toprule
		Method	& Year&A $\rightarrow$ W&D $\rightarrow$ W&W $\rightarrow$ D&A $\rightarrow$ D&D $\rightarrow$ A&W $\rightarrow$ A&Avg\\
		\midrule
		ResNet-50~\cite{he2016deep}&CVPR'2016&68.4$\pm$0.2&96.7$\pm$0.1&99.3$\pm$0.1&68.9$\pm$0.2&62.5$\pm$0.3&60.7$\pm$0.3&76.8\\
		DANN~\cite{ganin2016domain}&JMLR'2016&82.0$\pm$0.4&96.9$\pm$0.2&99.1$\pm$0.1&79.7$\pm$0.4&68.2$\pm$0.4&67.4$\pm$0.5&82.2\\
		CDAN+E~\cite{long2018conditional}&NIPS'2018&94.1$\pm$0.1&98.6$\pm$0.1&\textbf{100.0}$\pm$0.0&92.9$\pm$0.2&71.0$\pm$0.3&69.3$\pm$0.3&87.7\\
		SymNets~\cite{zhang2019domain}&CVPR'2019&90.8$\pm$0.1&98.8$\pm$0.3&\textbf{100.0}$\pm$0.0&93.9$\pm$0.5&74.6$\pm$0.6&72.5$\pm$0.5&88.4\\
		BSP+CDAN~\cite{chen2019transferability}&ICML'2019&93.3$\pm$0.2&98.2$\pm$0.2&\textbf{100.0}$\pm$0.0&93.0$\pm$0.2&73.6$\pm$0.3&72.6$\pm$0.3&88.5\\
		SPL~\cite{wang2020unsupervised}&AAAI'2020&92.7$\pm$0.0&98.1$\pm$0.0&99.8$\pm$0.0&93.7$\pm$0.0&76.4$\pm$0.0&76.9$\pm$0.0&89.6\\
		DMP~\cite{luo2020unsupervised}&PAMI'2020&93.0$\pm$0.3&99.0$\pm$0.1&\textbf{100.0}$\pm$0.0&91.0$\pm$0.4&71.4$\pm$0.2&70.2$\pm$0.2&87.4\\
		ATM~\cite{li2020maximum}&PAMI'2020&95.7$\pm$0.3&\textbf{99.3}$\pm$0.1&\textbf{100.0}$\pm$0.0&\textbf{96.4}$\pm$0.2&74.1$\pm$0.2&73.5$\pm$0.3&89.8\\
		DMRL~\cite{wu2020dual}&ECCV'2020&90.8$\pm$0.3&99.0$\pm$0.2&\textbf{100.0}$\pm$0.0&93.4$\pm$0.5&73.0$\pm$0.3&71.2$\pm$0.3&87.9\\
		MCC~\cite{jin2020minimum}&ECCV'2020&95.5$\pm$0.2&98.6$\pm$0.1&\textbf{100.0}$\pm$0.0&94.4$\pm$0.3&72.9$\pm$0.2&74.9$\pm$0.3&89.4\\
		Symnets-G~\cite{cui2020gradually}&CVPR'2020&93.8$\pm$0.4&98.8$\pm$0.2&\textbf{100.0}$\pm$0.0&\underline{96.1}$\pm$0.3&74.9$\pm$0.4&72.8$\pm$0.3&89.4\\
		SRDC*~\cite{tang2020unsupervised}&CVPR'2020&94.6$\pm$1.0&\underline{99.2}$\pm$0.5&\textbf{100.0}$\pm$0.0&92.6$\pm$0.6&\underline{78.1}$\pm$1.3&76.3$\pm$0.2&90.1\\
		RSDA-MSTN~\cite{gu2020spherical}&CVPR'2020&\textbf{96.1}$\pm$0.2&\textbf{99.3}$\pm$0.2&\textbf{100.0}$\pm$0.0&95.8$\pm$0.3&77.4$\pm$0.8&\textbf{78.9}$\pm$0.3&\textbf{91.1}\\
		\midrule
		SCAL&-&93.5$\pm$0.2&98.5$\pm$0.1&\textbf{100.0}$\pm$0.0&93.4$\pm$0.3&72.4$\pm$0.1&74.0$\pm$0.3&88.6\\
		SCAL+SPL&-&\underline{95.8}$\pm$0.3&\underline{99.2}$\pm$0.4&\textbf{100.0}$\pm$0.0&94.6$\pm$0.1&\underline{77.5}$\pm$0.2&76.0$\pm$0.2&\underline{90.5}\\
		\bottomrule
	\end{tabu}
\end{table*}

\begin{table}
	\centering
	\caption{Accuracy (\%) for Synthetic$\rightarrow$Real task of VisDA-2017 based on ResNet-101. (*) the implementation is based on the authors' code.}
	\label{tab:visda}
	\small
	\begin{tabular}{lc}
		\toprule
		Method&Synthetic$\rightarrow$Real\\
		\midrule
		ResNet-101~\cite{he2016deep}&52.4\\
		CDAN+E~\cite{long2018conditional}&71.7\\
		BSP+CDAN\cite{chen2019transferability}&75.9\\
		SPL*~\cite{wang2020unsupervised}&67.3\\
		DMP~\cite{luo2020unsupervised}&79.3\\
		ATM~\cite{li2020maximum}&75.1\\
		DMRL~\cite{wu2020dual}&75.5\\
		MCC~\cite{jin2020minimum}&78.8\\
		GVB-GD~\cite{cui2020gradually}&75.3\\
		RSDA-DANN~\cite{gu2020spherical}&75.8\\
		\midrule
		SCAL&80.1\\
		SCAL+SPL&\textbf{81.7}\\
		\bottomrule
	\end{tabular}
\end{table}

\subsection{Results}
We compare SCAL with state-of-the-art (SOTA) methods of DANN~\cite{ganin2016domain}, CDAN+E~\cite{long2018conditional}, SymNets~\cite{zhang2019domain}, BSP+CDAN~\cite{chen2019transferability}, SPL~\cite{wang2020unsupervised}, DMP~\cite{luo2020unsupervised}, ATM~\cite{li2020maximum}, AADA+CCN~\cite{yangmind}, DMRL~\cite{wu2020dual}, GVB-GD~\cite{cui2020gradually}, SRDC~\cite{tang2020unsupervised} and RSDA~\cite{gu2020spherical}. Most results of them are directly quoted from their original papers while a small portion of the results are obtained by running their online codes with the default training setting (denoted by *). To further investigate the effectiveness of the proposed framework, we also apply SPL~\cite{wang2020unsupervised}, a state-of-the-art pseudo-labeling method to SCAL (replacing the original clustering method with SPL).

\textbf{Office-Home.} As shown in~\cref{tab:office-Home}, we evaluate our methods on all the twelve transfer tasks of \emph{Office-Home}. We show that the proposed framework SCAL improves the average accuracy of CDAN~\cite{long2018conditional} by 2.5\%, showing the effectiveness of the proposed conditions. Compared to other domain adaptation methods, SCAL+SPL achieves a significant improvement as well.

\textbf{Office-31.}~\cref{tab:office-31} lists the average classification accuracies of the target samples on six transfer tasks of \emph{Office-31}. Results demonstrate that our SCAL+SPL achieves second-place performance with other domain adaptation methods. It is notable that SCAL+SPL boosts the average accuracy of the best method~\cite{gu2020spherical} on \emph{Office-31} by 1.1\% and 5.9\% on \emph{Office-Home} and \emph{VisDA-2017}, respectively.

\textbf{VisDA-2017.}~\cref{tab:visda} shows the results on \textbf{Synthetic}$\rightarrow$\textbf{Real} transfer task of \emph{VisDA-2017}. Following standard UDA setting~\cite{long2018conditional}, we use ResNet-101 as the feature extractor for VisDA-2017. The results of \cref{tab:visda} shows that SCAL is universal for different datasets.

\begin{table*}
	\caption{Ablation studies of our methods for Office-31 on individual components. (*) the implementation is based on the authors' code.}
	\label{tab:aba}
	\small
	\begin{tabu} to 0.99\textwidth{X[7l] X[2c] X[2c] X[2c] X[2c] X[2c] X[2c] X[c]}
		\toprule
		Method	&A $\rightarrow$ W&D $\rightarrow$ W&W $\rightarrow$ D&A $\rightarrow$ D&D $\rightarrow$ A&W $\rightarrow$ A&Avg\\
		\midrule
		\multicolumn{8}{l}{\emph{analysis of the conditions:}}\\
		\midrule
		Ours w/o conditions&82.0&96.9&99.1&79.7&68.2&67.4&82.2\\
		Ours with src.center&89.8&94.3&99.0&86.5&63.7&62.2&82.6\\
		Ours with src.KNN&92.6&98.1&\textbf{100.0}&86.3&67.0&67.3&85.2\\
		Ours with src.classifier&94.1&98.6&\textbf{100.0}&92.9&71.0&69.3&87.7\\
		Ours with tgt.k-means (last)&94.3&95.8&99.8&91.4&71.2&72.4&87.5\\
		SCAL&93.5&98.5&\textbf{100.0}&93.4&72.4&74.0&88.6\\
		SCAL+SPL&\textbf{95.8}&\textbf{99.2}&\textbf{100.0}&\textbf{94.6}&\textbf{77.5}&\textbf{76.0}&\textbf{90.5}\\
		\midrule
		\multicolumn{8}{l}{\emph{analysis of the differentiable approximation:}}\\
		\midrule
		Ours with non-differentiable&90.4&97.6&\textbf{100.0}&89.2&68.9&68.6&85.8\\		
		SCAL&93.5&98.5&\textbf{100.0}&93.4&72.4&74.0&88.6\\		
		\midrule
		\multicolumn{8}{l}{\emph{analysis of the base pipelines:}}\\
		\midrule
		JAN*&85.4&97.4&99.8&84.7&68.6&70.0& 84.3\\
		JAN with tgt.k-means (src.center)&93.0&97.9&99.8&88.0&68.6&68.2&85.9\\
		MCD*&79.5&97.6&\textbf{100.0}&81.1&61.2&61.5&80.1\\
		MCD with tgt.k-means (src.center)&89.8&97.0&98.6&89.2&76.3&73.8&87.4\\
		\bottomrule
	\end{tabu}
\end{table*}

\begin{figure*}[t]
	\centering
	\subfigure[Clustering]{
		\label{fig:clustering_result:a}
		\begin{minipage}[ht]{0.22\textwidth}
			\includegraphics[width = 1\columnwidth]{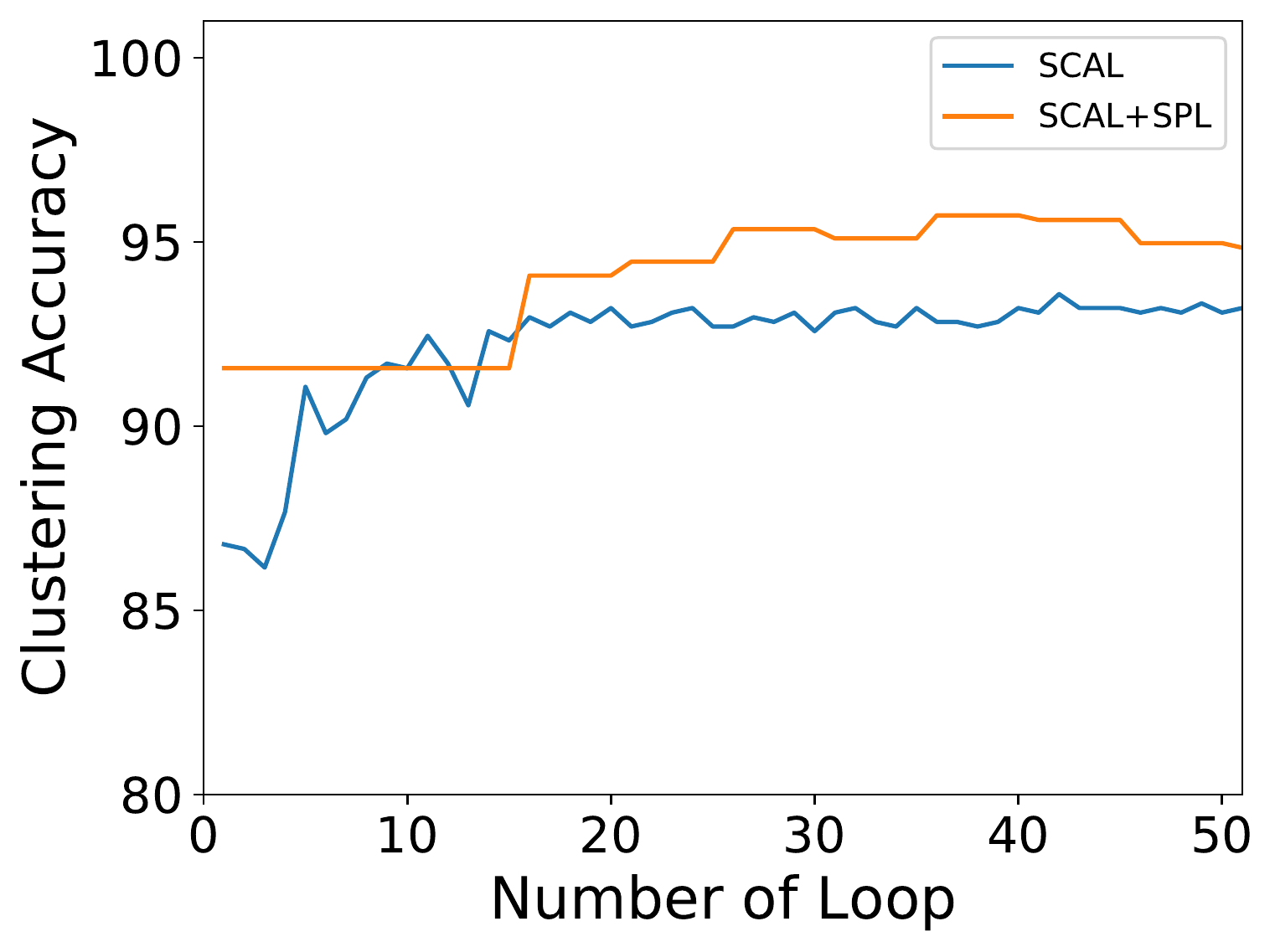}
		\end{minipage}	
	}	\subfigure[Visualization]{
		\label{fig:feature_visualization:b}
		\begin{minipage}[ht]{0.22\textwidth}
			\includegraphics[width = 1\columnwidth]{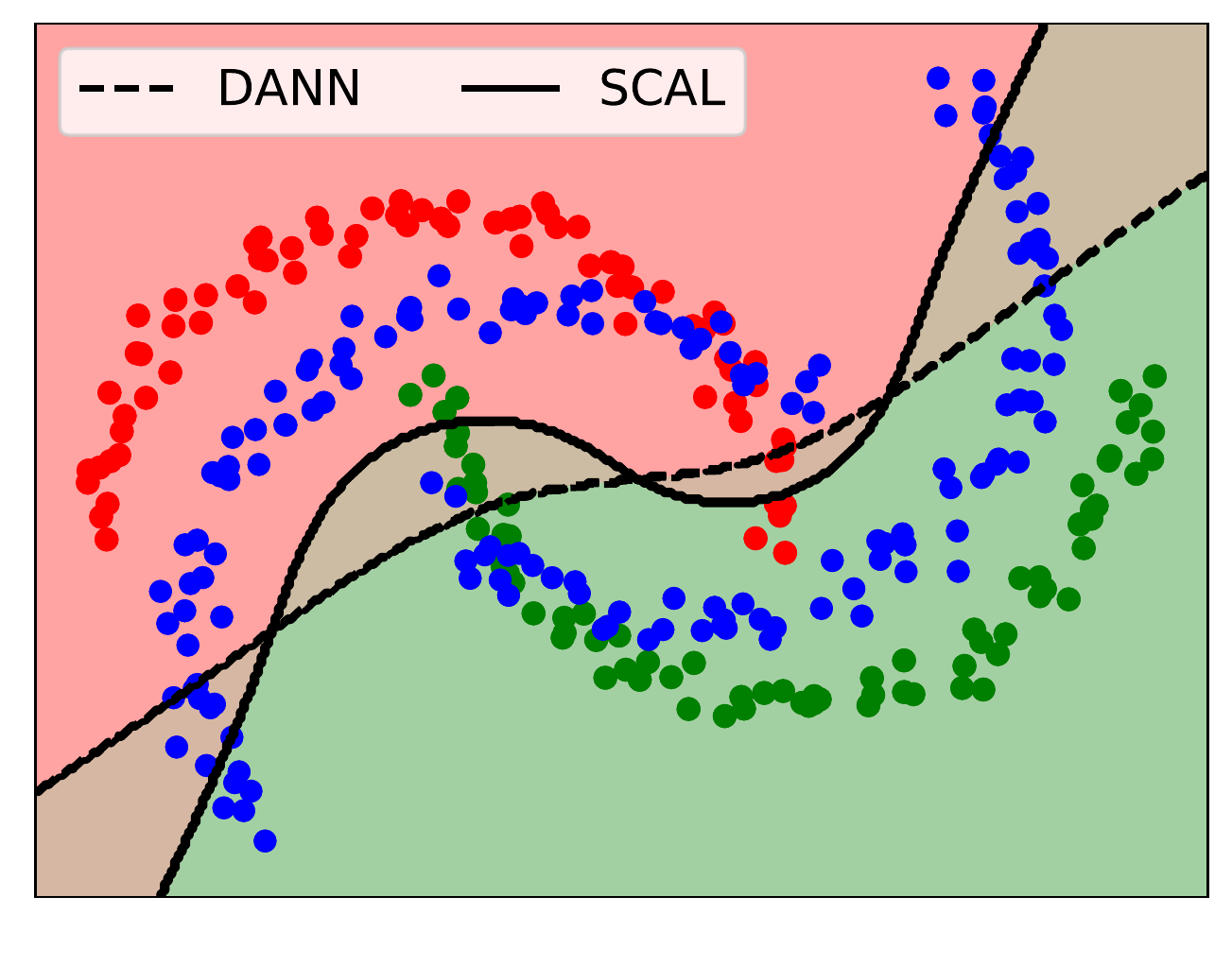}
		\end{minipage}	
	}
	\subfigure[Divergence]{
		\label{fig:feature_visualization:c}
		\begin{minipage}[ht]{0.22\textwidth}
			\includegraphics[width = 1\columnwidth]{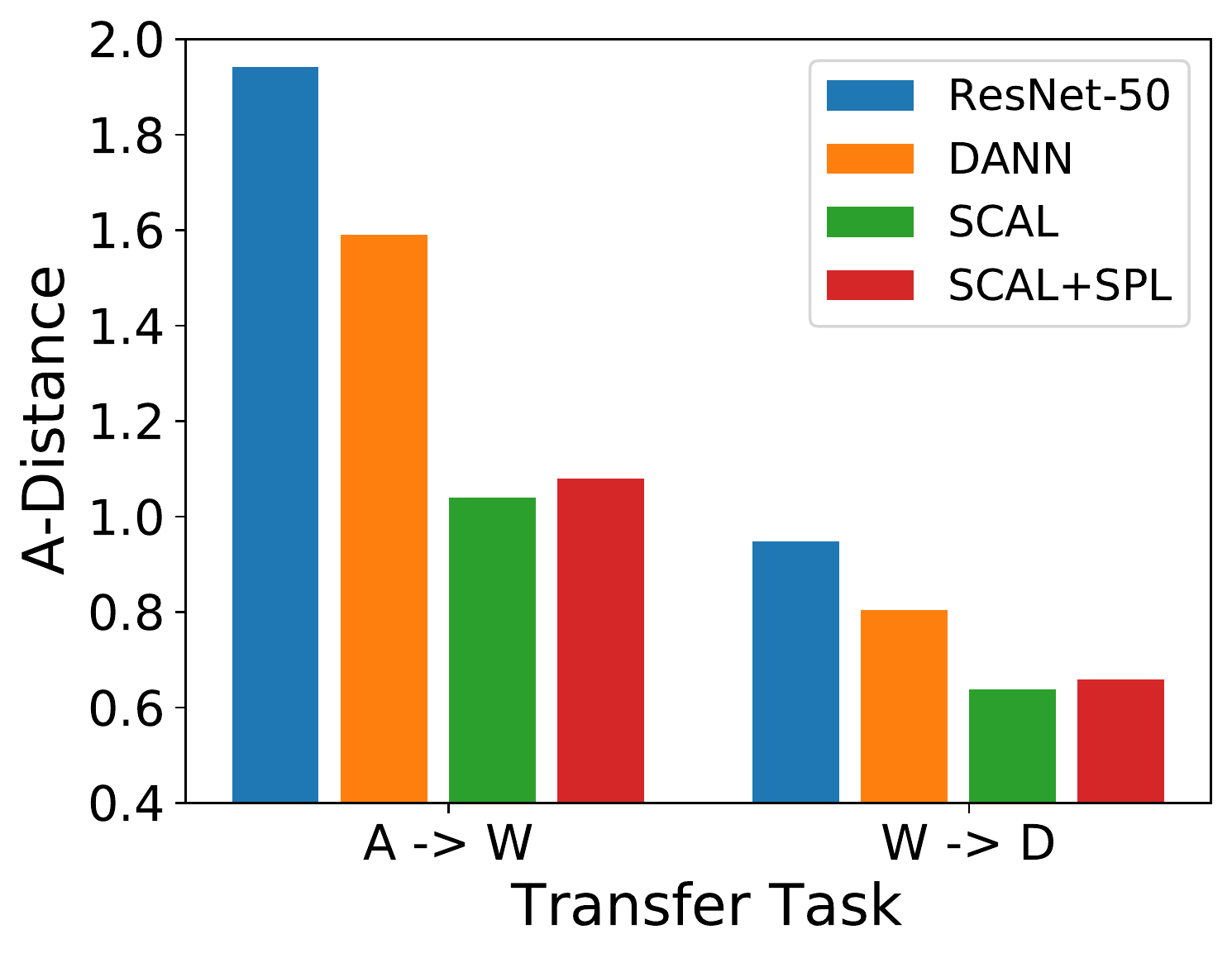}	
		\end{minipage}
	}
	\subfigure[Convergence]{
		\label{fig:feature_visualization:d}
		\begin{minipage}[ht]{0.22\textwidth}
			\includegraphics[width = 1\columnwidth]{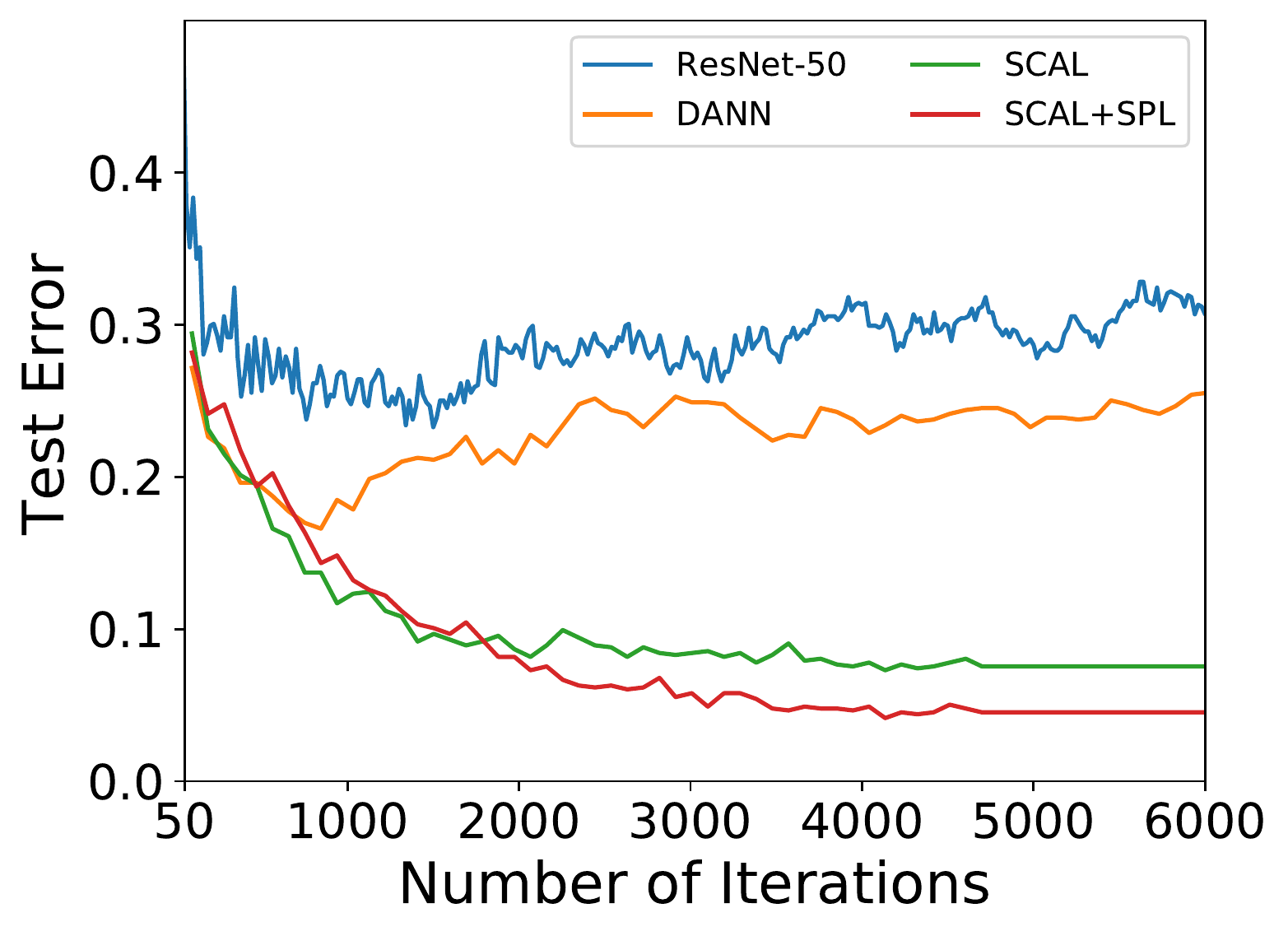}	
		\end{minipage}
	}
	\caption{Empirical analysis of clustering accuracy, visualization on the toy dataset, distribution divergence, and convergence performance. Please refer to the main text for how different experiments are defined.}
	\label{fig:feature_visualization}
\end{figure*}

\subsection{Empirical Analysis}\label{ea}

\textbf{Analysis of the Conditions.}
As shown in~\cref{tab:aba}, we examine the ablation experiments on \emph{Office-31} to investigate the effects of the conditions used in the adversarial adaptation method. Baseline ``Ours w/o conditions'' adopts the adversarial adaptation method without conditions. ``Ours with src.center'', ``Ours with src.KNN'' (K=1), and ``Ours with src.classifier'' treat the pseudo label obtained by the nearest source center, nearest K source features, and the predictions of source classifier as the conditions for the baseline. We can observe that the result is improved with the added conditions, showing the effectiveness of conditional distribution alignment. We then adopt the non-parametric clustering methods to model the target local structure. The difference between ``Ours with tgt.k-means (last)'' and SCAL is the center initialization way. The former utilizes the last clustering center to initialize the target center, while the latter utilizes the new source center to reinitialize the target center. In~\cref{tab:aba}, we can observe that the SCAL achieve better average accuracy than ``Ours with tgt.k-means (last)'' (+1.1\%), indicating the effectiveness of the source center initialization. We also employ SCAL+SPL to test the effect of our method when using a complex pseudo-labeling method, i.e., SPL~\cite{wang2020unsupervised}, which shows a better performance than SCAL by 1.9\%. These results of~\cref{tab:aba} demonstrate the importance of the conditions for adversarial learning. Different from previous methods, SCAL can integrate the effective non-differentiable clustering conditions into adversarial learning, thereby achieving a better performance. 

\begin{figure}[ht]
	\centering
	\subfigure[ResNet]{
		\label{fig:allcl:a}
		\begin{minipage}[ht]{0.30\columnwidth}
			\includegraphics[width = 1\columnwidth]{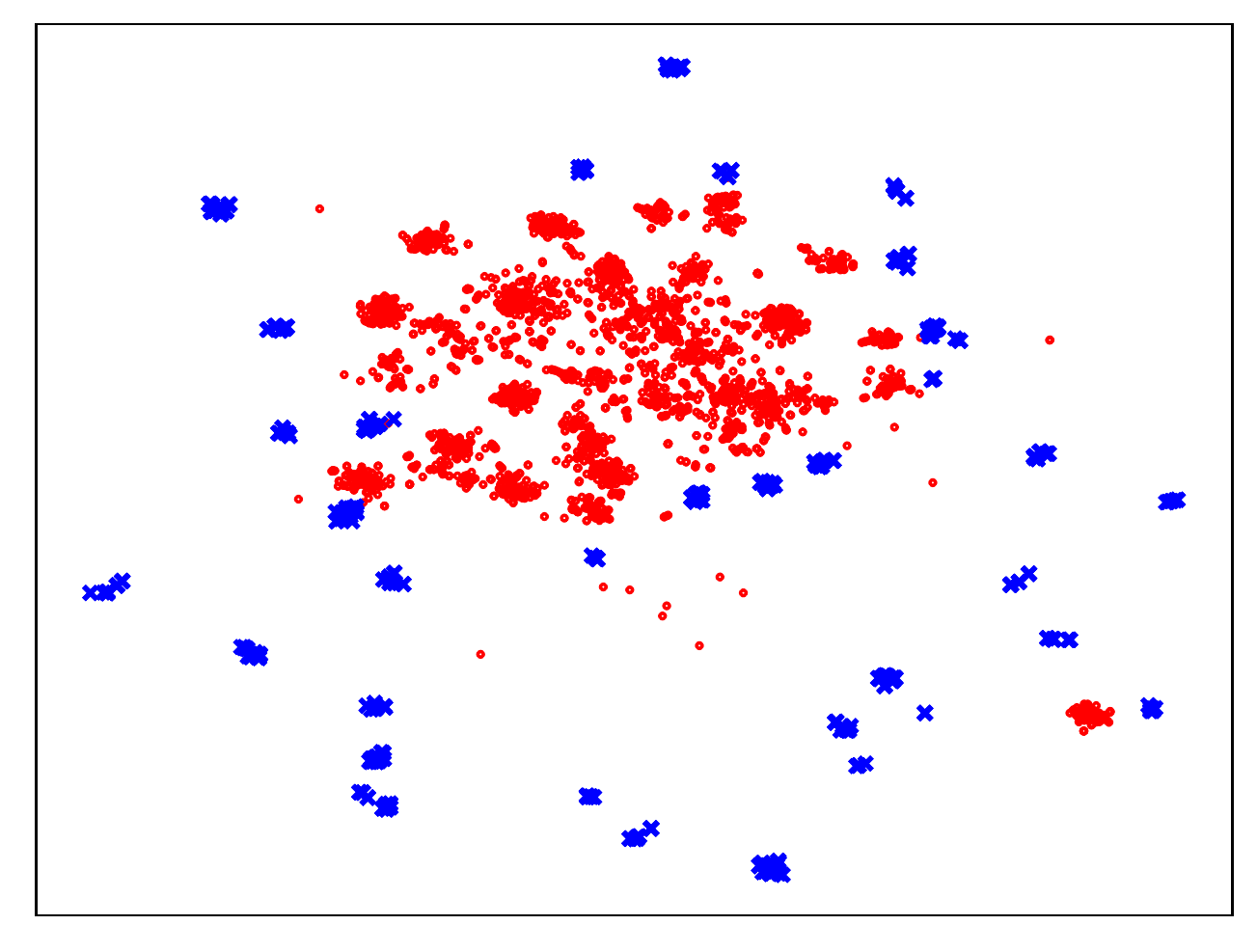}	
	\end{minipage}}
	\subfigure[DANN]{
		\label{fig:allcl:b}
		\begin{minipage}[ht]{0.30\columnwidth}
			\includegraphics[width = 1\columnwidth]{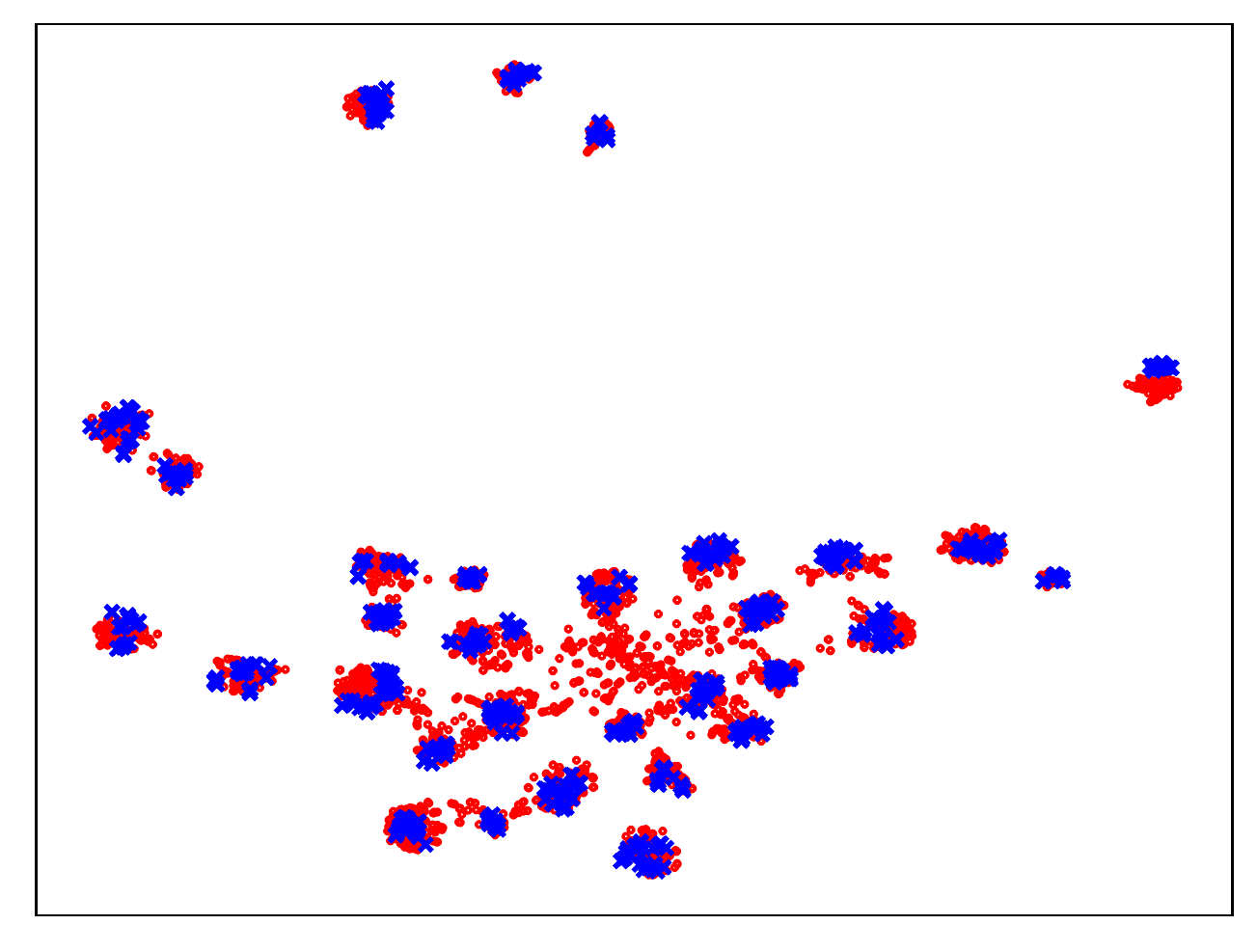}
		\end{minipage}	
	}
	\subfigure[Our Method]{
		\label{fig:allcl:c}
		\begin{minipage}[ht]{0.30\columnwidth}
			\includegraphics[width = 1\columnwidth]{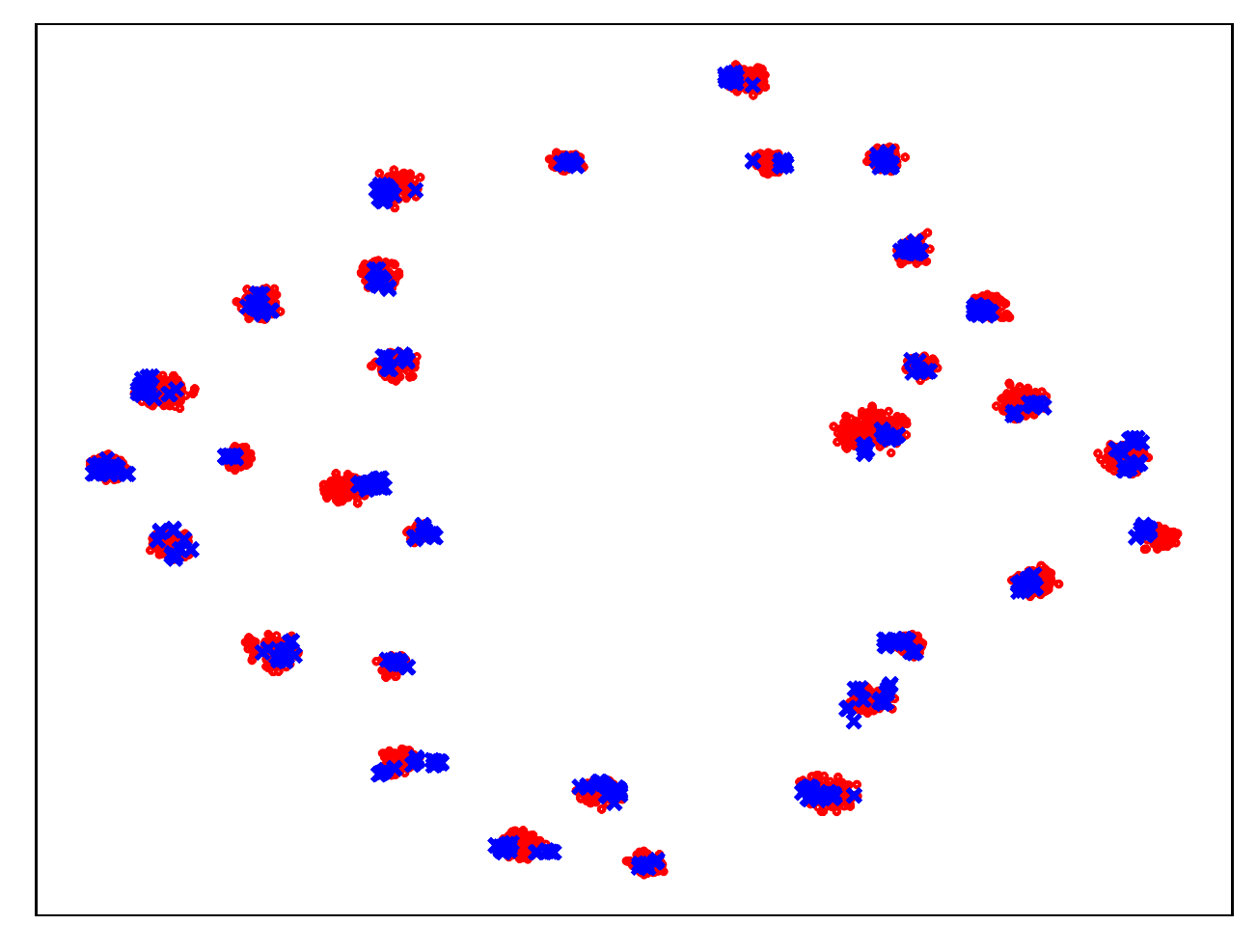}	
		\end{minipage}
	}
	\caption{Visualization of different methods on \textbf{W}$\rightarrow$\textbf{A} (all classes) of Office-31 (blue: \textbf{W}; red: \textbf{A}).}
	\label{fig:allcl}
\end{figure}

\begin{figure*}[ht]
	\centering
	\includegraphics[width = 1\textwidth]{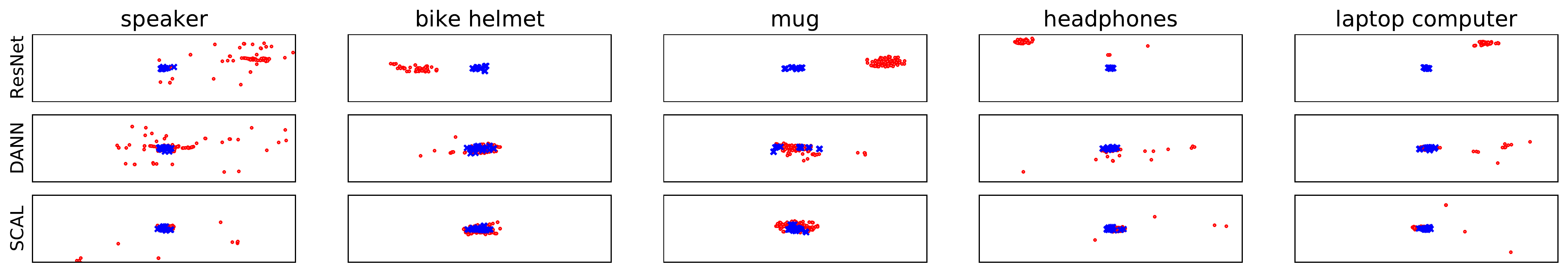}	
	\caption{Visualization of different methods on \textbf{W}$\rightarrow$\textbf{A} (1 classes) of Office-31 (blue: \textbf{W}; red: \textbf{A}).}
	\label{fig:disca2}
\end{figure*}

\textbf{Differentiable Approximation Analysis.} We test the strength of the differentiable approximation by comparing our method with the non-differentiable conditioned adversarial learning, i.e., ``Ours with non-differentiable''. This method uses the same clustering algorithm as SCAL, but simply uses the obtained pseudo-labels as conditions. SCAL boosts the average accuracy of ``Ours with non-differentiable'' by 2.8\%, which suggests the effectiveness of our surrogate classifier approximation.

\textbf{Baseline Analysis.} We further evaluate the performance of the structure-aware conditions by incorporating it into other types of UDA pipelines on \emph{Office-31}. MCD~\cite{saito2018maximum} is another line of the adversarial methods which do not use the domain classifier, and JAN~\cite{long2017deep} is a typical metric-based method. Similar to SCAL, ``JAN with tgt.k-means (src.center)'' and ``MCD with tgt.k-means (src.center)'' perform distribution alignment on the features conditioned on the outputs of the surrogate classifier. As shown in~\cref{tab:aba}, ``JAN with tgt.k-means (src.center)'' achieves better accuracy than JAN by 1.6\%, and ``MCD with tgt.k-means (src.center)'' boosts the average accuracy of MCD by 7.3\%. These results demonstrate the transferability of the proposed conditions.

\textbf{Clustering Accuracy Analysis.}
To deeply explore the advantages of SCAL over the pseudo-labeling methods, we report the accuracy curves of them in our framework. From \cref{fig:clustering_result:a}, we can observe that the clustering accuracies of SCAL and SCAL+SPL are both improved in the global adversarial alignment process. This supports the effectiveness of our framework.

\textbf{Visualization of Toy Dataset.}
We visualize the decision boundaries of different methods on inter twinning moons 2D problems in~\cref{fig:feature_visualization:b}. We follow the experimental setting of MCD~\cite{saito2018maximum}. The source samples of classes $0$ and $1$ are denoted by the red and green points. Blue points are target samples generated by rotating the source samples. The dark line and dashed line are the decision boundaries of DANN and SCAL. The result of SCAL is clearly better than DANN. 

\textbf{Divergence Analysis.}
The distribution divergence~\cite{ben2010theory} can be measured by the $\mathcal{A}$-distance~\cite{long2018conditional,ben2007analysis}. As shown in~\cref{fig:feature_visualization:c}, we observe that the $\mathcal{A}$-distance on ResNet-50 is larger than these adversarial methods, implying the efficacy of adversarial adaptation on reducing the domain gap. Besides, the $\mathcal{A}$-distance on SCAL is smaller than the $\mathcal{A}$-distance on DANN, manifesting that structure-conditioned adversarial learning can further reduce distribution divergence.

\textbf{Convergence Analysis.}
We testify the convergence performance of different methods on task \textbf{A} $\rightarrow$ \textbf{W} (31 classes) shown in~\cref{fig:feature_visualization:d}. Without the adversarial process, ResNet-50 enjoys faster convergence than other methods but the domain shift is preserved, thereby suffer an obvious accuracy drop. For the adversarial adaptation methods, SCAL and SCAL+SPL obtain consistent lower test error than DANN.

\textbf{Visualization of Feature Distribution.}
\cref{fig:10cl,fig:allcl} visualize the feature distribution under different granularity on task \textbf{W}$\rightarrow$\textbf{A} of \emph{Office-31} by t-SNE~\cite{maaten2008visualizing}. As shown in~\cref{fig:disca2}, the intra-class compactness exists in the feature distribution of the target domain, and the distribution alignment process has a potential risk of damaging it. This supports our motivation. And with the structure-aware conditions, our method can preserve the intra-class compactness and reduce the domain shift simultaneously. As shown in~\cref{fig:allcl}, compared to ResNet-50 and DANN, as expected, SCAL can preserve the intra-class compactness and reduce the domain shift as well.


\begin{table}[t]
	\centering
	\caption{Accuracy (\%) of SCAL with the different noise levels of clustering for A$\rightarrow$W task on Office-31.}
	\label{tab:noise}
	\small
	\begin{tabular}{lcc}
		\toprule
		Noise Level&w/o conditions&with noisy conditions\\
		\midrule
		0\%&82.0&93.5\\
		25\%&82.0&93.3\\
		50\%&82.0&91.6\\
		75\%&82.0&86.9\\
		100\%&82.0&74.7\\
		\bottomrule	
	\end{tabular}
\end{table}

\begin{table}[t]
	\centering
	\caption{Time (s) of SCAL variants in one epoch for different datasets based on ResNet-50.}
	\label{tab:time}
	\small
	\begin{tabu} to 0.99\columnwidth{X[3.5, l] X[c] X[c]}
		\toprule
		Task&Baseline&SCAL\\
		\midrule
		\textbf{A}$\rightarrow$\textbf{W} (Office-31)&48.3&55.4\\
		\textbf{Ar}$\rightarrow$\textbf{Cl} (Office-Home)&86.2&98.9\\
		\textbf{Synthetic}$\rightarrow$\textbf{Real} (VisDA-2017)&2260.3&2759.4\\
		\bottomrule	
	\end{tabu}
\end{table}

\textbf{Noise Sensitivity.}
We also investigate the sensitivity of SCAL with respect to different noise levels $nl$ on task \textbf{A}$\rightarrow$\textbf{W} for \emph{Office-31}. Specifically, we replace $nl\in\{0\%,25\%,50\%,75\%,100\%\}$ of the cluster-based pseudo-labels with the random labels and then integrate them as conditions into the adversarial process. Here we use the source classifier output as the final predictions. As shown in~\cref{tab:noise}, the accuracy of our method is reduced by less than 0.2\% when the noise level is $25\%$. Besides, even in the situation of a large noise level ($nl=75\%$), our method can boost the accuracy of "w/o conditions" by $4.9\%$.

\textbf{Time Cost Analysis.}
Besides the network training, our method needs to perform the clustering algorithm at each epoch. Therefore, we testify the time costs of the baseline method (i.e., DANN) and SCAL in one epoch on one Nvidia GeForce GTX 1080 Ti. As shown in~\cref{tab:time}, we need about 15\% additional time to perform the clustering algorithm, which is acceptable in the domain adaptation task.

\section{Conclusion}
In this paper, we have proposed an end-to-end structure-conditioned adversarial domain adaptation scheme, which aims to preserve the intra-class compactness through structure-aware conditions in the adversarial learning process. Specifically, the proposed scheme establishes local structures by the clustering algorithm and then incorporates the obtained structures into the adversarial learning process as structure-aware conditions for domain distribution alignment. The structure establishment and structure-conditioned adversarial learning have been iteratively performed. Experimental results have validated the effectiveness of the proposed method.


%




\ifCLASSOPTIONcaptionsoff
  \newpage
\fi



\bibliographystyle{IEEEtran}
\bibliography{IEEEabrv,IEEEtran}
\end{document}